\documentclass[twoside]{article}
\usepackage[accepted]{aistats2023}
\usepackage{color, amsfonts, enumitem, graphicx, amssymb, amsthm, bbm, algorithm, algorithmic, pgfplotstable, booktabs, subfig, mathtools, pifont, dsfont}
\usepackage[round]{natbib}
\usepackage[hidelinks]{hyperref}
\makeatletter
\def\munderbar#1{\underline{\sbox\tw@{$#1$}\dp\tw@\z@\box\tw@}}
\makeatother

\newtheorem{remark}{Remark}

 \newtheorem{definition}{Definition}[section]

\newtheorem{lemma}[definition]{Lemma}

\definecolor{cadmiumgreen}{rgb}{0.0, 0.42, 0.24}

\newcommand{\be}{\begin{equation}}
\newcommand{\ee}{\end{equation}}
\newcommand{\bea}{\begin{equation*}\begin{aligned}}
\newcommand{\eea}{\end{aligned}\end{equation*}}
\newcommand{\ds}{\displaystyle}

\newcommand{\R}{\mathbb{R}}

\newcommand{\Max}{\max\limits_}
\newcommand{\Min}{\min\limits_}

\DeclareMathOperator*{\argmax}{argmax}
\DeclareMathOperator*{\argmin}{argmin}

\newcommand{\wh}{\widehat}
\newcommand{\mc}{\mathcal}

\DeclareMathOperator{\st}{s.t.}

\newcommand{\PSD}{\mathbb{S}_{+}} 
\newcommand{\Let}{\triangleq}
\newcommand{\opt}{^\star}

\newcommand{\cmark}{\ding{51}}
\newcommand{\xmark}{\ding{55}}

\usepackage{url}

\begin{document}


\twocolumn[

\aistatstitle{Feasible Recourse Plan via Diverse Interpolation}

\aistatsauthor{ Duy Nguyen \And Ngoc Bui \And  Viet Anh Nguyen }

\aistatsaddress{ VinAI Research, Vietnam \And  VinAI Research, Vietnam \And The Chinese University of Hong Kong } ]


\begin{abstract}
     Explaining algorithmic decisions and recommending actionable feedback is increasingly important for machine learning applications. Recently, significant efforts have been invested in finding a diverse set of recourses to cover the wide spectrum of users' preferences. However, existing works often neglect the requirement that the recourses should be close to the data manifold; hence, the constructed recourses might be implausible and unsatisfying to users. To address these issues, we propose a novel approach that explicitly directs the diverse set of actionable recourses towards the data manifold. We first find a diverse set of prototypes in the favorable class that balances the trade-off between diversity and proximity. We demonstrate two specific methods to find these prototypes: either by finding the maximum a posteriori estimate of a determinantal point process or by solving a quadratic binary program. To ensure the actionability constraints, we construct an actionability graph in which the nodes represent the training samples and the edges indicate the feasible action between two instances. We then find a feasible path to each prototype, and this path demonstrates the feasible actions for each recourse in the plan. The experimental results show that our method produces a set of recourses that are close to the data manifold while delivering a better cost-diversity trade-off than existing approaches.
\end{abstract}




\section{Introduction} \label{sec:intro}
Algorithmic recourse is an emerging method to explain machine learning (ML) models by suggesting how to alter the predictive outcome of any given instance~\citep{ref:karimi2020survey, ref:verma2020counterfactual, ref:stepin2021survey, ref:pawelczyk2021carla}. For example, when a bank deploys an ML model to filter credit loan applicants, it is desirable to provide the denied applicants with certain explanations on why such a decision was made. One possible explanation comes in the form of a directive recommendation such as \textit{``you need to do A to be granted a loan''}, where A is a concrete action to be implemented. This type of explanation is called an algorithmic \textit{recourse} and a set of multiple recourses is called a \textit{recourse plan}. The recourse plan provides a reasoning for the decision and suggests actions that need to perform if the users want to reapply in the future. Algorithmic recourse is becoming a powerful tool to enhance the reliability of the ML model and the engagement of the users, especially in the consequential domains such as loan approvals~\citep{ref:siddiqi2012credit}, university admission~\citep{ref:waters2014grade}, and job hiring~\citep{ref:ajunwa2016hiring}, to name a few. Algorithmic recourse~\citep{ref:ustun2019actionable} is also known in the literature of interpretable machine learning as counterfactual explanation~\citep{ref:wachter2017counterfactual, ref:bui2022counterfactual} or contrastive explanation~\citep{ref:karimi2020survey}.

In practice, many criteria should be considered when constructing a recourse plan. First, each recourse in the plan should be valid in the sense that the recourse should flip the unfavorable outcome of the predictive model. Second, the actions should be relatively small to alleviate the human efforts in implementing the recourses. Third, the recourses should be diverse to cover a wide spectrum of users' preferences. Finally, the recourses should be attainable: the recommended recourses should be actionable for the users who receive them~\citep{ref:ustun2019actionable}. A recourse plan that takes into account these four criteria is called feasible.

Several techniques have been invested in constructing a diverse recourse plan for a given classifier~\citep{ref:russell2019efficient, ref:mothilal2020explaining, ref:dandl2020multi,  ref:bui2022counterfactual}. \citet{ref:russell2019efficient} proposes a mixed-integer programming method to generate a counterfactual plan for a linear classifier, in which the diversity is imposed using a rule-based approach. \citet{ref:dandl2020multi} propose a model-agnostic approach using a multi-objective evolutionary algorithm to construct a diverse recourse plan. Alternatively, a recourse plan can also be found by using iterative methods to minimize the weighted sum of three terms representing validity,
proximity, and diversity~\citep{ref:mothilal2020explaining}.

To promote the actionability of the constructed recourses, existing approaches often use constraints to restrict the space of admissible recourses in the optimization problem~\citep{ref:russell2019efficient, ref:dandl2020multi, ref:mothilal2020explaining}. The optimization problem is then solved by common constraint programming (CP) solvers or projected gradient descent algorithms. However, these approaches often require expert knowledge to design suitable actionability constraints. Furthermore, they often consider the input features independently; thus the constructed recourses might be implausible and unsatisfying to the users. For example, the recourse might suggest the user has an income of $10000$ with a job that has an income of $1000$ on average. To address this issue, one might use the structural causal model (SCM) and find the recourse through the minimal intervention on the SCM graph~\citep{ref:karimi2020algorithmic, ref:karimi2021algorithmic}. However, this approach generates only a single recourse and requires access to the SCM, which is rarely available in practice~\citep{ref:verma2020counterfactual}. 

FACE~\citep{ref:poyiadzi2020face} is another method for constructing sequential and actionable recourses.
To capture actionability constraints, FACE constructs a graph where the nodes represent the training samples and the edges indicate if there is a feasible action that can transform one node into another. FACE then finds the prototypes in a high-density region of the training data with favorable prediction and recommends sequential and actionable recourses as the shortest paths from the input to the prototypes.

Other approaches consider the data manifold when constructing recourse plans~\citep{ref:pawelczyk2020learning, ref:joshi2019towards, ref:looveren2021interpretable}. The recourse is considered attainable if it is close to the manifold of the training dataset. \citet{ref:joshi2019towards} and \citet{ref:pawelczyk2020learning} model the data manifold by learning a variational autoencoder (VAE) for the distribution of the training data. \citet{ref:looveren2021interpretable} use an example in the desired class, namely \textit{prototype}, to guide the constructed recourse toward the data manifold. However, these works only consider a single recourse.

\textbf{Problem Statement.} We consider a binary classification setting with a $p$-dimensional covariate and a binary $\{0, 1\}$ response, where $0$ represents a negative/unfavorable outcome and $1$ represents a positive/favorable outcome. In real-world applications, the negative outcome may correspond to decisions such as ``loan denied" or ``application rejected", while the positive outcome may correspond to ``loan approved" or ``application accepted". We are given a classifier $\mc C: \R^p \to \{0, 1\}$, and we assume that we have access to $N$ data samples $\wh x_i \in \R^p$, $i = 1, \ldots, N$ which has positive predicted outcome: $\mc C(\wh x_i) = 1$. Given an input $x_0 \in \R^p$ which receives a negative predicted outcome by the model, i.e., $\mc C(x_0) = 0$, we aim to provide a menu, or a plan, consisting $K$ recourses $\{x_k^r\}_{k = 1, \ldots, K}$ for $x_0$ that balances multiple criteria such as cost, validity, diversity, and adherence to the data manifold. These criteria can be translated as follows: The cost criterion implies that $x_k^r$ should be close to the input $x_0$, where closeness is measured using a certain distance function. The validity requires that $\mc C(x_k^r) = 1$, while the diversity requires that $\{x_k^r\}$ should be sufficiently different from each other. The adherence to the data manifold here requires that the recourses $x_k^r$ should be relatively close to the aforementioned samples $\wh x_i$.

\textbf{Our approach and contributions.} Our recourse plan generator consists of two stages:
\begin{enumerate}[label=(\roman*), leftmargin=7mm]
    \item Find a set of $K$ prototypes $\{x_k \}_{k = 1, \ldots, K}$ from the available dataset $\{ \wh x_i\}$.
    \item For each $k$, find the corresponding recourse $x_k^{r}$ via an interpolation from $x_0$ to $x^k$.
\end{enumerate}
Our generator is based on the premise that the interpolation mechanism preserves the desirable properties of the prototypes, including diversity and proximity. Under this premise, if $\{x_k\}$ are diverse then the recourses $\{x_k^r\}$ are also diverse. Moreover, if $x_k$ is close to the input $x_0$, it is also likely that the interpolation will lead to a recourse $x_k^r$ that is close to $x_0$. Thus, it is imperative to induce diversity and proximity in Step (i) above when we find the $K$ prototypes. Nevertheless, we acknowledge that there is currently no general consensus on the ``correct" definition of the diversity for a recourse plan, and different users may prefer different metrics to measure the diversity of a plan.

Guided by this thinking, we introduce two novel methods to find a set of $K$ prototypes that balance the trade-off between the diversity and proximity from $x_0$. In Section~\ref{sec:det}, we formulate a proximity-based determinantal point process and find the prototypes through a maximum a posteriori estimate using a greedy and local search heuristic. In Section~\ref{sec:quad}, we take a different viewpoint on the trade-off, and we formulate a binary quadratic program (BQP) to select the prototypes. We then use an eigen-approximate hierarchy to approximate this BQP, which can be solved efficiently by either a best-response or a dual-ascent method. This hierarchy serves as a screening tool to identify potential samples and subsequently reduce the size of the BQP.

Our method relies on the interpolation module: given an input $x_0$ and a prototype $x_k$, the interpolation traverses on a path joining these two endpoints and returns a point $x_k^r$ on this path that is closest to $x_0$ and has positive prediction $\mc C(x_k^r) = 1$. There are many possible ways to form a path emanating from $x_0$ to connect with $x_k$. If no actionability constraints are imposed, we can use a linear interpolation, which corresponds to tracing a straight line joining $x_0$ and~$x_k$. If actionability is required, then we can employ a graph representation similar to FACE~\citep{ref:poyiadzi2020face} and use the shortest path to traverse from $x_0$ to $x_k$.

\begin{figure}[!ht]
    \centering
    \vspace{-5mm}
    \includegraphics[width=\linewidth]{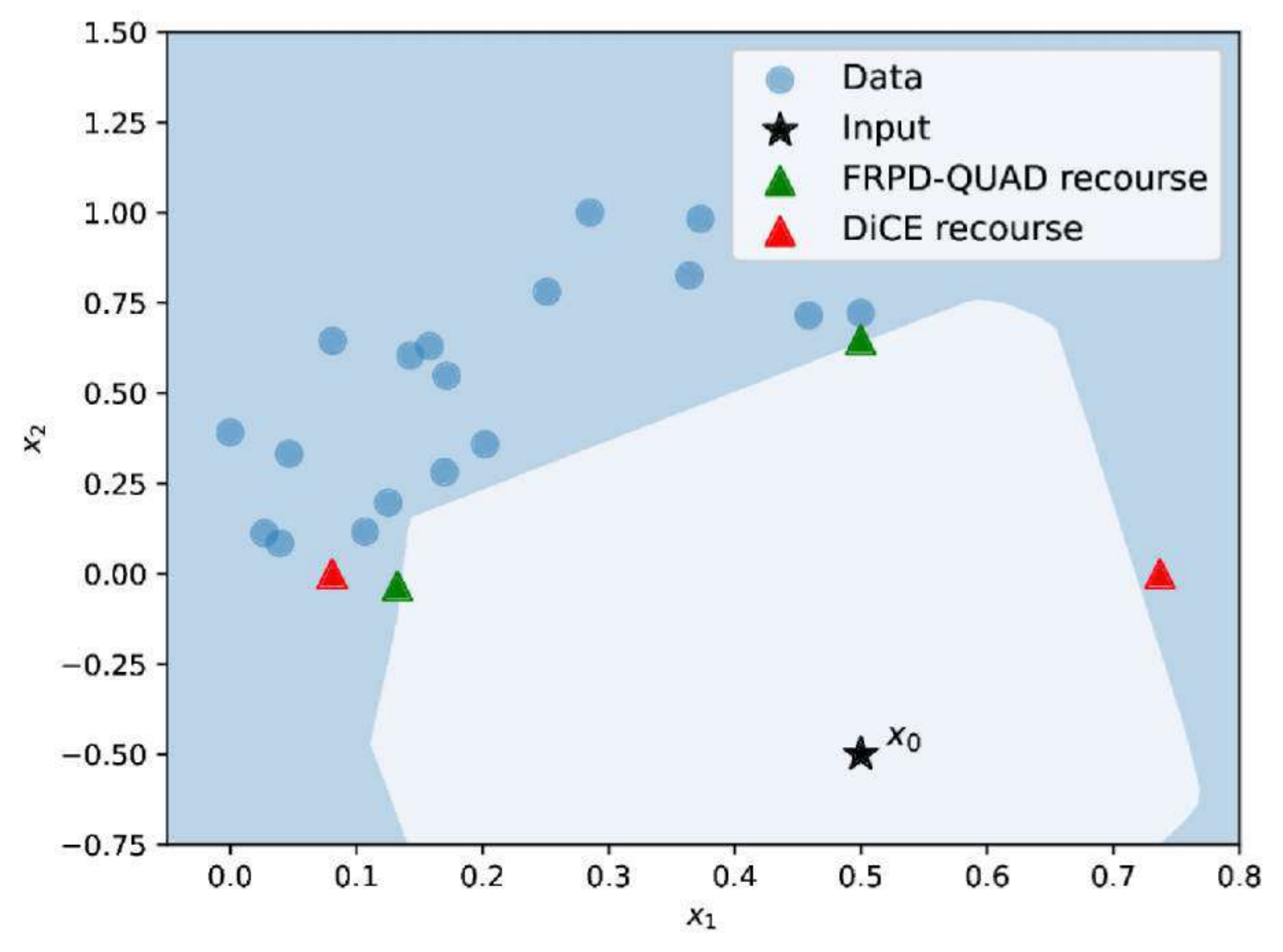}
    \vspace{-7mm}
\caption{The right, red triangle is suggested by DiCE and it fails to adhere to the data manifold.}
    \label{fig:plot_2D}
    \vspace{-3mm}
\end{figure}

\begin{figure}[!ht]
    \centering
    \vspace{-4mm}
    \includegraphics[width=\linewidth]{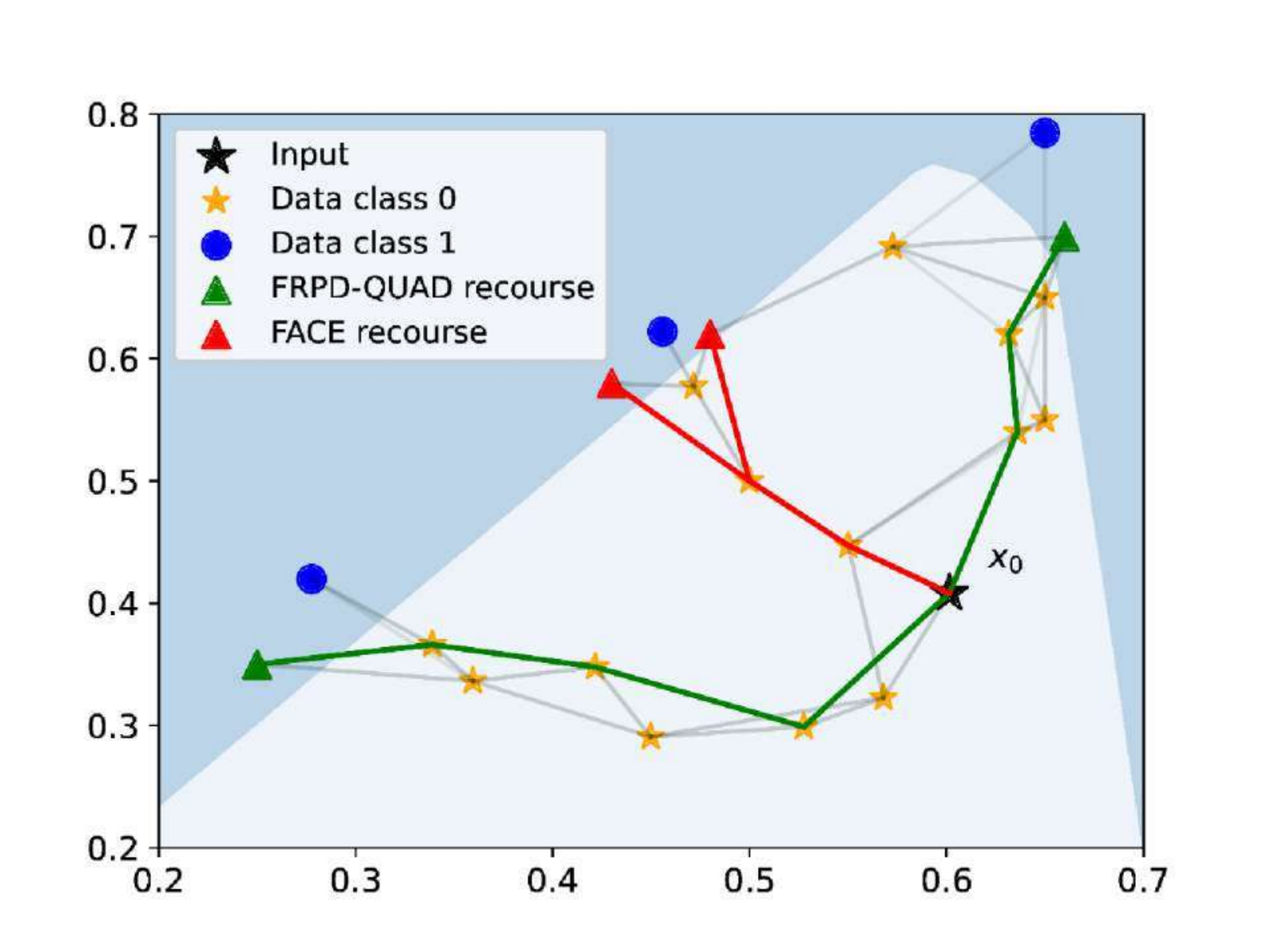}
    \vspace{-7mm}
    \caption{The red triangles are suggested by FACE and they are not sufficiently diverse.}
    \label{fig:frpd_face}
    \vspace{-3mm}
\end{figure}

Our proposed scheme has several advantages vis-\`{a}-vis the state-of-the-art methods for generating diverse recourse plans, notably the DiCE~\citep{ref:mothilal2020explaining}, and methods for generating sequential and actionable recourses, such as FACE~\citep{ref:poyiadzi2020face}:
    \begin{enumerate}[leftmargin = 5mm]
        \item First, by identifying the prototypes and then conducting an interpolation, we guarantee that the recourse $x_k^r$ is directed towards the samples with the favorable predicted value. Notice that DiCE does not assume to have access to the samples $\{ \wh x_i\}$ from the favorable class, thus the recourses generated from DICE may not flow towards $\{ \wh x_i\}$. As a consequence, our recourses will adhere better to the data manifold than the recourses of DiCE, see Figure~\ref{fig:plot_2D}. In addition, by using an appropriate method of interpolation, we guarantee that our recourses are valid, in a sense that $\mc C(x_k^r) = 1$. Because DiCE is a gradient-based method, there is no guarantee that the recourses generated by DiCE are uniformly valid.
        \item Second, our formulation takes the diversity measured with respect to the input $x_0$. More specifically, we measure the diversity of prototypes by the diversity of the normalized direction vectors pointing from the input $x_0$ to the samples $\wh x_i$. In doing so, the diversity will be adjusted to the coordinate of the input vector $x_0$. In contrast, the approach from DiCE measures the diversity directly from the samples $\wh x_i$ and is not adjusted to $x_0$.
        \item Third, by using a prototype selection method and an actionability graph, we ensure that the recourses are actionable and diverse. In contrast, the FACE method does not consider the diversity criteria explicitly, thus it may fail to produce a sequential recourse plan that is diverse, see Figure~\ref{fig:frpd_face} for an example.
    \end{enumerate}
    
Figure~\ref{fig:plot_2D} shows a two-dimensional example in which the DiCE method recommends a recourse that does not adhere to the data manifold. The positively predicted samples are drawn as blue circles, and the input $x_0$ is represented by a black star. The background colors represent the positive and negative classification regions. The DiCE recourses (red triangles) are diverse, but one recourse fails to be close to the data manifold (the recourse drawn on the right). The reason is that DiCE is a gradient-based method that does not take the data into account. On the contrary, our recourses found by solving a BQP (green triangles) are correctly directed towards the data manifold. 

Figure~\ref{fig:frpd_face} shows a two-dimensional example of recourses generated under the actionability constraints. The grey lines represent all edges of the actionability graph. The green and red lines represent the recourse actions for input $x_0$ for our method and FACE, respectively. The FACE recourses are not diverse because FACE does not directly optimize this criterion.

\textbf{Additional notations.} Given $x_0$ and a sample $\wh x_i$, we set the \textit{normalized} direction vector $a_i$ as \be \label{eq:direction}
        a_i \Let (\wh x_i - x_0)/\|\wh x_i - x_0 \|_2 \in \R^p
    \ee
    so that each $a_i$ has unit $l_2$-norm. The matrix $A = [a_1, \ldots, a_N] \in \R^{p \times N}$ collects all the direction vectors, and we compute the (cosine) similarity matrix as $S = A^\top A$, where we can write each element in $S$ by $S_{ij} = a_i^\top a_j$. The element $S_{ij}$ is also closely related to the cosine angle between the direction vectors of $\wh x_i$ and $\wh x_j$. The space of $N$-by-$N$ symmetric positive definite matrices is denoted by $\PSD^N$. For any vector $z$, $\|z \|_0$ counts the number of non-zero elements in $z$. We write $I$ to denote the identity matrix of the appropriate dimension.

\section{Prototype Selection via Determinantal Point Processes} \label{sec:det}

In this section, we construct the set of diverse prototypes using a maximum a posteriori (MAP) estimate of a determinantal point process (DPP).  For the purpose of this paper, the choice of the prototypes should balance the trade-off between diversity and proximity with respect to the input $x_0$. We first provide a brief introduction to DPP, a more thorough introduction can be found in Appendix~\ref{sec:dpp}. Then, we construct a proximity-based DPP and discuss how to find its MAP estimate to select the prototypes.

DPPs is a family of probabilistic measures that arise from the field of quantum physics: it is particularly useful to model the repulsive behavior of Fermion particles~\citep{ref:macchi1975coincidence}. Grounded by its probabilistic nature, DPPs have been applied in various machine learning tasks~\citep{ref:kulesza2012determinantal, ref:affandi2014learning, ref:urschel2017learning}, ranging from text and video summarization~\citep{ref:lin2021learning, ref:cho2019multi, ref:gong2014diverse} to recommendation systems~\citep{ref:chen2018fast, ref:wilhelm2018practical, ref:gartrell2017low}. 

\begin{definition}[$L$-ensemble DPP] \label{def:dpp-L}
    Given a positive semidefinite $N$-by-$N$ matrix $L \in \PSD^N$, an $L$-ensemble DPP is a distribution over all $2^N$ index subsets $J \subseteq \{1, \ldots, N\}$ such that
\[\mathrm{Prob}(J) = \det(L_J)/ \det(I + L),\]
where $L_J$ denotes the $|J|$-by-$|J|$ submatrix of $L$ with rows and columns indexed by $J$.
\end{definition}


We need to construct a DPP that aids the selection of $K$ diverse prototypes from $N$ data samples that have favorable predicted outcomes. Because an $L$-ensemble DPP can be identified via a matrix $L \in \PSD^N$, we focus on choosing an appropriate $L$ that can balance between two criteria: the diversity of the prototypes and the distance of the prototypes to the input $x_0$. Towards this end, we impose an additional parameter $\theta \in [0, 1]$ to capture this trade-off, and the matrix $L^\theta$ is defined as
\[
    L^\theta = \theta  S + (1-\theta)  D,
\]
where $S$ is a similarity matrix among favorable data samples and $D$ is a diagonal matrix capturing the locality structure around $x_0$. One possible choice of $S$ is the cosine similarity matrix $ S = A^\top A \in \PSD^N$, and one possible choice of $D$ is the diagonal matrix with
\[
 D_{ii} = \exp(-\mathrm{dist}(x_0, \wh x_i)^2/h^2) \quad \forall i = 1, \ldots, N,
\]
where $h > 0$ is the bandwidth and $\mathrm{dist}$ is a distance function. Notice that because both $S$ and $D$ chosen as such are positive semidefinite, $L^\theta$ is also positive semidefinite. We then find the $K$ diverse prototypes from the training data by solving the following problem
\be \label{eq:det}
    \max \left\{ \det ( L_z^\theta) ~:~ z \in \{0, 1\}^N,~ \| z \|_0= K \right\},
\ee
where $L_z^\theta$ is a submatrix of $ L^\theta$ restricted to rows and columns indexed by the one-components of $z$. It is well-known that the solution to problem~\eqref{eq:det} coincides with the MAP estimate of the DPP with a cardinality constraint~\citep{ref:kulesza2012determinantal}. 

Let us now consider the two extremes: If $\theta = 0$, then $L^0 = D$, it is now easy to see that the optimal solution $z\opt$ of~\eqref{eq:det} will have non-zeros elements corresponding to the maximum values of $D_{ii}$. Thus, $z\opt$ translates to choosing $K$ data samples that are closest to $x_0$, measured by the distance function $\mathrm{dist}$. On the other extreme, if $\theta = 1$, then $L^1 = S$, and we recover the canonical setting of DPP \textit{without} the proximity constraints. The resulting optimal solution of~\eqref{eq:det} in this case will simply promote the diversity of the prototypes.


We now switch gears to discuss the solution procedure to solve~\eqref{eq:det}. First, we highlight that problem~\eqref{eq:det} is a submodular maximization problem since the log-probability function in DPP $\log \det( L_z)$ is a submodular function~\citep{ref:gillenwater2012near}. Further, this problem is well-known to be NP-hard~\citep{ref:kulesza2012determinantal}, and thus it is notoriously challenging to solve~\eqref{eq:det} to optimality. In addition, in real-world applications of the machine learning model, we expect the number of samples $N$ to be large. We thus resort to popular heuristics in order to find a good solution to~\eqref{eq:det} in a high-dimensional setting with low solution time. A common greedy algorithm to solve the MAP estimation problem~\eqref{eq:det} is to iteratively find at each incumbent set of prototypes $z$ an index $j$ by
\[ 
    j = \argmax_{i: z_i = 0}~\log\det(L_{z \vee e_i }) - \log \det(L_z),
\]
where $\vee$ returns the element-wise maximum between two vectors and $e_i$ is the vector of zeros with the $i$-th element being one. The algorithm then adds $j$ to the set of prototypes until reaching the cardinality constraint, i.e., when $\|z\|_0 = K$. In doing so, each $j$ guarantees to maximize the marginal gain to the incumbent set. An efficient implementation of this greedy algorithm that costs $\mc O(K^2N)$ time for each inference is provided in~\cite{ref:chen2018fast}.

\textbf{Local search for MAP inference. } The greedy algorithm achieves an approximation ratio of $\mc O(\frac{1}{k!})$~\citep{ref:civril2009selecting}. To improve its performance, we introduce in Appendix~\ref{sec:ls} a simple $2$-neighborhood local search that switches one element from the incumbent set with one element from the complementary set.


\section{Prototype Selection via Quadratic Binary Programming} \label{sec:quad}

We now describe a second approach to finding the set of prototypes that can balance the diversity and proximity trade-off. In stark contrast to the DPP method proposed in the previous section, this second method relies solely on formulating a binary quadratic program in an intuitive manner and then utilizing optimization techniques in order to scale up the problem to the high-dimensional setting. 

\textbf{Formulation.} Given the normalized direction vectors $a_i$ defined as in~\eqref{eq:direction}, we can measure the anti-diversity of a set of prototypes by a quadratic form
\[
    \mathrm{Anti\text{-}Diversity}(z) \Let \sum\nolimits_{i,j} a_i^\top a_j z_i z_j
\]
that simply sums up the pairwise inner product of direction vectors. It is easy to see that if two prototypes $i$ and $j$ have similar directions $a_i$ and $a_j$, then they are not diverse, which is further represented by the fact that their inner product $a_i^\top a_j$ are close to one. Thus, higher values of $\mathrm{Anti\text{-}Diversity}(z)$ indicate that the solution $z$ is not collectively diverse.

Given a distance function $\mathrm{dist}$, we can now formulate a quadratic program whose objective function captures the diversity and proximity trade-off. For a weight parameter $\vartheta \in [0, 1]$, we define the following optimization problem
\[
\begin{array}{cll}
\min & \vartheta  \mathrm{Anti\text{-}Diversity}(z) +  (1-\vartheta)\sum\nolimits_{i}  \mathrm{dist}(\wh x_i, x_0) z_i\\
\st & z \in \{0, 1\}^N,~ \| z \|_0 = K.
\end{array}
\]
In the matrix notations, we have the equivalent form
\be\label{eq:cosine2}
\begin{array}{cll}
        \min & \vartheta z^\top S z + (1-\vartheta) d^\top z \\
        \st & z \in \{0, 1\}^N,~ \| z \|_0 = K,
\end{array}
\ee
where $S \in \PSD^N$ is the similarity matrix, and $d \in \R_+^N$ is the vector of distances from $x_0$ with $d_i = \mathrm{dist}(\wh x_i, x_0)$. If $\vartheta = 0$, then problem~\eqref{eq:cosine2} only takes the proximity into account and it collapses into a linear binary program. Its optimal solution can be identified using a greedy argument; indeed, the optimal solution $z\opt$, in this case, has $z_i\opt = 1$ if the sample $\wh x_i$ is one of the $K$ nearest samples from $x_0$. When $\vartheta = 1$, problem~\eqref{eq:cosine2} stresses only on minimizing the anti-diversity, or equivalently, maximizing the diversity of the prototypes. Because $S \in \PSD^N$, problem~\eqref{eq:cosine2} has a convex quadratic objective function, and it can be solved by state-of-the-art solvers such as CPLEX~\citep{ref:cplex2009v12}, GUROBI~\citep{ref:gurobi2021gurobi} or Mosek~\citep{ref:mosek}. Nevertheless, it has $N$ binary variables and is not easy to solve if $N$ is large. Next, we explore approximation methods that find good quality solutions to~\eqref{eq:cosine2} with low computing overhead.

\textbf{Eigen-approximate hierarchy.} We now delineate our approach to solve~\eqref{eq:cosine2}, which is inspired by recent advances in using the eigen-approximations to solve (convex) quadratic binary programs~\citep{ref:vreugdenhil2021principal}. Suppose that $S$ admits the following eigendecomposition $S = \sum_{m=1}^N \sigma_m v_m v_m^\top$, where $\sigma_m$ are nonnegative eigenvalues and $v_m$ forms an orthogonal basis of $\R^N$. Without any loss of generality, we suppose that $\sigma_m$ are sorted in decreasing order, that is, $\sigma_1 \ge \sigma_2 \ge \ldots, \ge \sigma_N \ge 0$. Next, we approximate $S$ using its top-$M$ eigenspace approximation, i.e., $S \approx S_M \Let \sum_{m=1}^M \sigma_m v_m v_m^\top$, where we assume that $\sigma_M > 0$. In this case, $S_M$ is a strictly positive definite matrix.
Following~\cite{ref:vreugdenhil2021principal}, the resulting top-$M$ eigen-approximation of problem~\eqref{eq:cosine2} is
    \be\label{eq:cosineK}
    \begin{array}{cll}
        \min & (1-\vartheta) d^\top z + \vartheta z^\top S_M z \\
        \st & z \in \{0, 1\}^N,~ \| z \|_0 = K,
    \end{array}
    \ee
where we emphasize that the objective function of~\eqref{eq:cosineK} involves the matrix $S_M$.  The eigen-approximation is particularly useful thanks to its following min-max representation.
    
\begin{lemma}[Min-max equivalence] \label{lemma:minmax}
    Suppose that $\vartheta \in (0, 1]$. Problem~\eqref{eq:cosineK} is equivalent to the min-max problem
    \be \label{eq:minmax}
    \begin{array}{ccl}
        \Min{z \in \{0, 1\}^N, \|z\|_0 = K} & \Max{\gamma \in \R^M} & \mc L(z, \gamma)
     \end{array}
    \ee
    where $\mc L$ admits the following form:   
    \begin{multline}
        \mc L(z, \gamma) = ((1 - \vartheta) d - 2 \sum\nolimits_{m=1}^M \gamma_m v_m)^\top z \\
        - \sum\nolimits_{m=1}^M \gamma_m^2/(\vartheta \sigma_m).        
    \end{multline}
       
\end{lemma}
The proof of Lemma~\ref{lemma:minmax} is included in Appendix~\ref{sec:app-proof}. The advantage of the min-max formulation is that its objective function is a linear function of the binary variables $z$. This linearity is beneficial because it leads to an analytical optimal solution in the variable $z$ for any fixed value of $\gamma$.

Fix any vector $\gamma \in \R^M$, the optimization problem $\min_{z \in \{0,1\}^N,~\|z\|_0 = K}~\mc L(z, \gamma) $ admits the solution \be \label{eq:z-opt} z\opt(\gamma) = \mathds{1}_N (\mathcal I(\gamma)),\ee where the function $\mathds{1}_N(\mathcal I(\gamma))$ returns a binary vector whose elements in the index set $\mc I(\gamma)$ is one, and the set $\mc I(\gamma)$ is
    \[
        \mc I(\gamma) = \left\{\begin{array}{l} i \in \{1, \ldots, N\}: 
            i \text{ is an index of the $k$-smallest} \\
            \text{elements of vector } (1 - \vartheta) d - 2 \sum_{m=1}^M \gamma_m v_m
        \end{array}
        \right\}.
    \]
    Similarly, for any $z \in \{0, 1\}^N$, the problem over the $\gamma$ variable $\max_{\gamma \in \R^M}~\mc L(z,\gamma)$ admits the solution 
    \be \label{eq:gamma-opt} \gamma\opt(z) = \begin{pmatrix}
            -\vartheta \sigma_1 v_1^\top z ; \cdots ;
            -\vartheta \sigma_M v_M^\top z
    \end{pmatrix} \in \R^M,
    \ee
    which can be verified by the first-order optimality condition. Equipped with this information and inspired by the algorithms in~\cite{ref:vreugdenhil2021principal}, we deploy two different algorithms to screen potential solutions for~\eqref{eq:cosineK}:

\begin{algorithm}[H]
	\caption{Best response iterations}
	\label{alg:br}
	\begin{algorithmic}
		\STATE {\bfseries Input:} Eigenvalues $\sigma_1, \ldots, \sigma_M > 0$ with eigenvectors $v_1, \ldots, v_M$, distance vector $d$, weight $\vartheta$, limit iteration $T$
		\STATE {\bfseries Initialization:} 
		Set $z_{0} \leftarrow 0, \gamma_0 \leftarrow 0$
        \FOR{$t =0, \ldots, T-1$}
            \STATE \vspace{-5mm}
            \begin{align*}
                \hspace{-2mm}\gamma_{t+1} \!\leftarrow\!\! \begin{pmatrix}
                    -\vartheta \sigma_1 v_1^\top z_{t} \\
                    \vdots \\
                    -\vartheta \sigma_M v_M^\top z_{t}
                \end{pmatrix}~;~
                z_{t+1}\!\leftarrow\!\mathds{1}_N (\mathcal I(\gamma_{t+1}))
            \end{align*} 
        \ENDFOR
		\STATE{\bfseries Output:} $z_1, \ldots, z_T$
	\end{algorithmic}
\end{algorithm}

\textbf{Algorithm 1 (Best response iterations). } This algorithm leverages the best response functions in~\eqref{eq:z-opt} and~\eqref{eq:gamma-opt} to generate an alternating scheme. The iterations proceed by fixing an incumbent solution in $z$ to find the corresponding optimal solution in the $\gamma$ variable, then switch the role of these two variables to find the optimal solution in the $z$ variable corresponding to the incumbent solution of $\gamma$. The pseudocode is presented in Algorithm~\ref{alg:br}.

\textbf{Algorithm 2 (Dual ascent iterations). } This algorithm proceeds with the dual form of the min-max problem~\eqref{eq:minmax}, obtained by interchanging the max and min operators:
    \be \label{eq:dualprogram}
        \begin{array}{ccl}
            \Max{\gamma \in \R^M} & \Min{z \in \{0, 1\}^N, \|z\|_0 = K} & \mc L(\gamma, z)
        \end{array}
    \ee
    where $\mc L$ admits the following form:
    \begin{multline}
        \mc L(\gamma, z) = \ds (1 - \vartheta) d^\top z - \sum\nolimits_{m=1}^M \gamma_m^2/(\vartheta \sigma_m) \\
        - 2 \sum\nolimits_{m=1}^M \gamma_m v_m^\top z.        
    \end{multline}

    It is well-known that the optimal value of problem~\eqref{eq:dualprogram} constitutes a lower bound on that of problem~\eqref{eq:minmax}. The dual ascent iterations leverage the solution~\eqref{eq:z-opt} to solve the inner problem of~\eqref{eq:dualprogram}, then it computes the gradient in the outer $\gamma$ variable and takes a gradient ascent step with diminishing step sizes. The pseudocode is given in Algorithm~\ref{alg:dual}.
    
\begin{algorithm}[H]
	\caption{Dual program iterations}
	\label{alg:dual}
	\begin{algorithmic}
		\STATE {\bfseries Input:} Similar to Algorithm~\ref{alg:br}
		\STATE {\bfseries Parameters:}  $\lambda$  (Default $\lambda=0.1$)
		\STATE {\bfseries Initialization:} 
		Set $z_{0} \leftarrow 0$,  $\gamma_0 \leftarrow 0$
        \FOR{$t =0, \ldots, T-1$}
            \STATE \vspace{-5mm}
            \begin{align*}
                \hspace{-3mm}\gamma_{t+1} &\leftarrow \gamma_{t} - \frac{2\lambda}{\sqrt    {t + 1}} \begin{pmatrix}
            \frac{ \gamma_1}{\vartheta \sigma_1} + v_1^\top z_t \\ 
            \vdots \\
            \frac{ \gamma_m}{\vartheta \sigma_m} + v_m^\top z_t
        \end{pmatrix}
                \\
                \hspace{-3mm}z_{t+1} &\leftarrow \mathds{1}_N (\mathcal I(\gamma_{t+1})),
            \end{align*} 
        \ENDFOR
		\STATE{\bfseries Output:} $z_1, \ldots, z_T$
	\end{algorithmic}
\end{algorithm}

\textbf{Screening and dimensionality reduction.} Both the best response and the dual ascent iterations can serve as screening tools to reduce the dimension of the original problem~\eqref{eq:cosine2}. Let $\tau < T$ be an integer. Intuitively, we look at the last $\tau+1$ iterate solutions in the variable $z$ and select the indices of non-zero among them. This set of indices will identify the samples that are likely to constitute the optimal solution for the eigen-approximate problem~\eqref{eq:cosineK}. Formally, this set can be found as $\mc Z = \{ i \in [N]: (z_{T- \tau} \vee \ldots \vee z_{T})_i > 0\}$. The set $\mc Z$ will be injected to problem~\eqref{eq:cosine2} to form a new problem
\be \label{eq:cosine-final}
\begin{array}{cll}
        \min & \vartheta z^\top S z + (1-\vartheta) d^\top z \\ 
        \st & z \in \{0, 1\}^N,~ \| z \|_0 = K,~z_i = 0 ~\forall i \not\in \mc Z, 
\end{array}
\ee
where all indices of $z$ which are not in the set $\mc Z$ are pre-set to zero. This is equivalent to reducing the number of binary variables, and problem~\eqref{eq:cosine-final} becomes more amenable to commercial solvers. The solution $z\opt$ of~\eqref{eq:cosine-final} determines the set of $K$ prototypes for interpolation.

\section{Interpolation Schemes} \label{sec:network}

Sections~\ref{sec:det} and~\ref{sec:quad} provide $K$ diverse prototypes to guide users for implementation. However, these recourses might be too costly and too far from the actual effort needed to flip the model prediction. A potential remedy is to use \textit{linear interpolations} between the input instance $x_0$ and the prototypes $\{x_k\}$ to find the recourses on the decision boundary~\citep{ref:vlassopoulos2020explaining}. In this linear interpolation scheme, the recourses are the intersections between the decision boundary and the line segments joining the input $x_0$ and the prototypes. 

The linear interpolation scheme will provide the plan with the lowest cost that preserves the diversity of $K$ prototypes. However, it is hard to impose actionability constraints into the linear interpolation; thus the recommended recourses could become implausible and unsatisfactory for users. It is notable that in Euclidean spaces, linear interpolation can guarantee that the diversity of recourses is preserved if we use the Euclidean distance in the construction of the similarity matrices for both the DPP (Section~\ref{sec:det}) and the QUAD (Section~\ref{sec:quad}) methods. Notice that state-of-the-art methods in the recourse literature, such as~\citet{ref:poyiadzi2020face} and \citet{ref:mothilal2020explaining} rely on the Euclidean space setup, and we are not aware of any work that generates diverse recourses in a non-Euclidean space.

To promote actionable and sequential recourses, we leverage the ideas from FACE~\citep{ref:poyiadzi2020face} to build a directed graph $\mc G = (\mc V, \mc E)$: each node $v \in \mc V$ corresponds to a sample in the training dataset, and an edge $(u, v) \in \mc E$ represents an actionable transition from node $u$ to node $v$. The actionable transition should respect the cost threshold constraints as well as immutable feature constraints which are prescribed in each dataset. A sequential recourse is a directed path from the input instance $x_0$ to a prototype, each transition in the path is considered an action. We then provide a diverse set of sequential recourses as the shortest paths connecting $x_0$ to $\{x_k\}$. Compared to FACE, our prototypes chosen from Sections~\ref{sec:det} and~\ref{sec:quad} are already diverse, hence the paths suggested by our method are more likely to be diverse than those suggested by FACE. Further details about sequential recourse are relegated to Appendix~\ref{sec:app-exp}.

\section{Numerical Experiments} 
\label{sec:expt}
We analyze our prototype selection variants: Greedy DPP (FRPD-DPP-GR), Local search DPP (FRPD-DPP-LS), and Quadratic (FRPD-QUAD) under non-actionability and actionability environments. For the non-actionability comparison, we use linear interpolation to find the recourses. In this line of comparison, we compare our approaches against DiCE~\citep{ref:mothilal2020explaining}, which is the most popular method to generate diverse recourses. For the actionability comparisons, we find the recourses based on the actionability graph, then compare our methods against FACE~\citep{ref:poyiadzi2020face}, which is the state-of-the-art method to generate actionable and sequential recourses. Appendix~\ref{sec:app-exp} reports details about experimental settings and additional numerical results. 


\begin{figure*}[!ht]
    \centering
    \includegraphics[width=\linewidth]{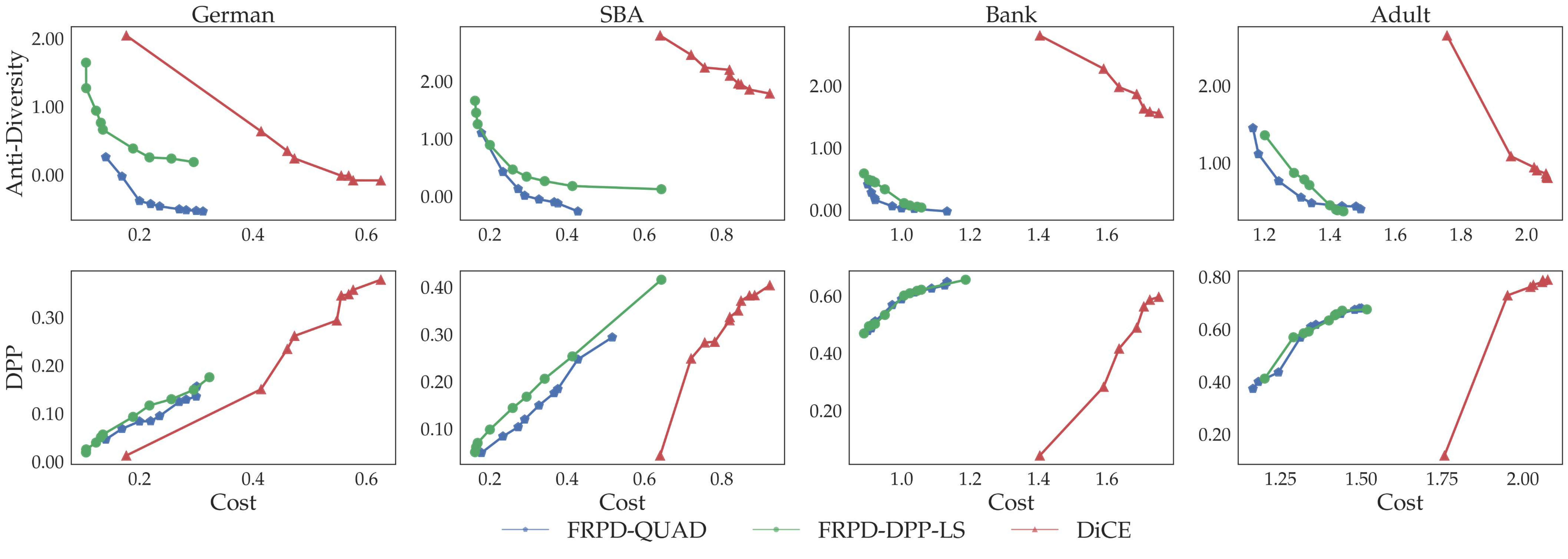}
    \caption{Pareto frontier of the cost vs.~Anti-Diversity (top) and cost vs.~DPP (bottom) trade-off on four real-world datasets.}
    \label{fig:mlp_pareto}
\end{figure*}

\begin{table*}[!ht]
    \centering
    \caption{Benchmark of Cost, Validity, Anti-Diversity, DPP and Manifold-Distance between competing methods with\textit{out} actionability constraints. Details on experiment setting are provided in Appendix~\ref{sec:app-exp}.}
    \label{tab:manifold_dist}
    \footnotesize
    \begin{filecontents*}{mlp_updated.csv}
dataset,method,cost,valid,diversity,dpp,manifold_dist
German,DiCE,0.41 $\pm$ 0.34,\textbf{1.00} $\pm$ 0.00,0.71 $\pm$ 0.74,0.14 $\pm$ 0.09,\textbf{0.15} $\pm$ 0.11
,FRPD-QUAD,0.30 $\pm$ 0.19,\textbf{1.00} $\pm$ 0.00,\textbf{-0.53} $\pm$ 0.70,0.14 $\pm$ 0.12,0.19 $\pm$ 0.19
,FRPD-DPP-GR,0.31 $\pm$ 0.22,\textbf{1.00} $\pm$ 0.00,0.17 $\pm$ 0.20,\textbf{0.16} $\pm$ 0.14,0.27 $\pm$ 0.23
,FRPD-DPP-LS,\textbf{0.30} $\pm$ 0.18,\textbf{1.00} $\pm$ 0.00,0.18 $\pm$ 0.20,0.15 $\pm$ 0.12,0.30 $\pm$ 0.23
SBA,DiCE,0.71 $\pm$ 0.22,0.95 $\pm$ 0.22,2.49 $\pm$ 0.61,0.23 $\pm$ 0.12,0.86 $\pm$ 0.24
,FRPD-QUAD,0.45 $\pm$ 0.30,\textbf{1.00} $\pm$ 0.00,\textbf{-0.17} $\pm$ 0.60,0.23 $\pm$ 0.19,\textbf{0.23} $\pm$ 0.16
,FRPD-DPP-GR,\textbf{0.41} $\pm$ 0.33,\textbf{1.00} $\pm$ 0.00,0.18 $\pm$ 0.21,0.24 $\pm$ 0.23,0.26 $\pm$ 0.19
,FRPD-DPP-LS,0.41 $\pm$ 0.33,\textbf{1.00} $\pm$ 0.00,0.18 $\pm$ 0.19,\textbf{0.25} $\pm$ 0.23,0.27 $\pm$ 0.19
Bank,DiCE,1.62 $\pm$ 0.32,\textbf{1.00} $\pm$ 0.00,2.14 $\pm$ 0.64,0.34 $\pm$ 0.26,0.48 $\pm$ 0.37
,FRPD-QUAD,1.09 $\pm$ 0.21,\textbf{1.00} $\pm$ 0.00,\textbf{0.03} $\pm$ 0.26,\textbf{0.63} $\pm$ 0.11,0.51 $\pm$ 0.20
,FRPD-DPP-GR,\textbf{1.06} $\pm$ 0.21,\textbf{1.00} $\pm$ 0.00,0.04 $\pm$ 0.11,0.62 $\pm$ 0.11,\textbf{0.43} $\pm$ 0.20
,FRPD-DPP-LS,\textbf{1.06} $\pm$ 0.21,\textbf{1.00} $\pm$ 0.00,0.04 $\pm$ 0.11,0.62 $\pm$ 0.11,\textbf{0.43} $\pm$ 0.20
Adult,DiCE,1.95 $\pm$ 0.37,\textbf{1.00} $\pm$ 0.00,1.08 $\pm$ 0.55,\textbf{0.73} $\pm$ 0.13,1.54 $\pm$ 0.31
,FRPD-QUAD,1.48 $\pm$ 0.58,\textbf{1.00} $\pm$ 0.00,0.43 $\pm$ 0.33,0.67 $\pm$ 0.18,\textbf{0.60} $\pm$ 0.32
,FRPD-DPP-GR,\textbf{1.44} $\pm$ 0.61,\textbf{1.00} $\pm$ 0.00,\textbf{0.37} $\pm$ 0.30,0.67 $\pm$ 0.19,0.66 $\pm$ 0.36
,FRPD-DPP-LS,\textbf{1.44} $\pm$ 0.61,\textbf{1.00} $\pm$ 0.00,\textbf{0.37} $\pm$ 0.30,0.67 $\pm$ 0.19,0.66 $\pm$ 0.36
\end{filecontents*}
\pgfplotstabletypeset[
        col sep=comma,
        string type,
        every head row/.style={before row=\toprule,after row=\midrule},
        every row no 3/.style={after row=\midrule},
        every row no 7/.style={after row=\midrule},
        every row no 11/.style={after row=\midrule},
        every last row/.style={after row=\bottomrule},
        columns/dataset/.style={column name=Dataset, column type={l}},
        columns/method/.style={column name=Methods, column type={l}},
        columns/cost/.style={column name=Cost, column type={c}},
        columns/valid/.style={column name=Validity, column type={c}},
        columns/diversity/.style={column name=Anti Diversity, column type={c}},
        columns/dpp/.style={column name=DPP, column type={c}},
        columns/manifold_dist/.style={column name=Manifold Distance, column type={c}},
    ]{mlp_updated.csv}
\end{table*}

\begin{table*}[htb]
    \centering
    \caption{Benchmark of Cost, Validity, Anti-Diversity, and DPP between competing methods with actionability constraints on four real-world datasets. Details on experiment setting are provided in Appendix~\ref{sec:app-exp}.}
    \label{tab:action}
    \footnotesize
    \begin{filecontents*}{mlp_action.csv}
dataset,method,cost,valid,diversity,dpp
German,FACE,\textbf{0.24} $\pm$ 0.17,\textbf{1.00} $\pm$ 0.00,2.18 $\pm$ 0.77,0.04 $\pm$ 0.04
,FRPD-QUAD,0.55 $\pm$ 0.24,\textbf{1.00} $\pm$ 0.00,\textbf{-0.13} $\pm$ 0.24,0.21 $\pm$ 0.07
,FRPD-DPP-GR,0.42 $\pm$ 0.20,\textbf{1.00} $\pm$ 0.00,0.20 $\pm$ 0.23,0.21 $\pm$ 0.05
,FRPD-DPP-LS,0.42 $\pm$ 0.20,\textbf{1.00} $\pm$ 0.00,0.22 $\pm$ 0.23,\textbf{0.23} $\pm$ 0.07
SBA,FACE,\textbf{0.25} $\pm$ 0.20,\textbf{1.00} $\pm$ 0.00,2.08 $\pm$ 0.78,0.02 $\pm$ 0.03
,FRPD-QUAD,0.55 $\pm$ 0.25,\textbf{1.00} $\pm$ 0.00,\textbf{-0.04} $\pm$ 0.41,0.17 $\pm$ 0.09
,FRPD-DPP-GR,0.52 $\pm$ 0.25,\textbf{1.00} $\pm$ 0.00,0.41 $\pm$ 0.25,\textbf{0.21} $\pm$ 0.03
,FRPD-DPP-LS,0.52 $\pm$ 0.25,\textbf{1.00} $\pm$ 0.00,0.41 $\pm$ 0.25,\textbf{0.21} $\pm$ 0.03
Bank,FACE,\textbf{1.29} $\pm$ 0.29,\textbf{1.00} $\pm$ 0.00,1.42 $\pm$ 0.86,0.31 $\pm$ 0.27
,FRPD-QUAD,1.55 $\pm$ 0.44,\textbf{1.00} $\pm$ 0.00,\textbf{0.06} $\pm$ 0.12,0.69 $\pm$ 0.16
,FRPD-DPP-GR,1.52 $\pm$ 0.42,\textbf{1.00} $\pm$ 0.00,0.16 $\pm$ 0.26,\textbf{0.74} $\pm$ 0.04
,FRPD-DPP-LS,1.52 $\pm$ 0.42,\textbf{1.00} $\pm$ 0.00,0.16 $\pm$ 0.26,\textbf{0.74} $\pm$ 0.04
Adult,FACE,\textbf{1.33} $\pm$ 0.36,\textbf{1.00} $\pm$ 0.00,2.66 $\pm$ 0.36,0.22 $\pm$ 0.27
,FRPD-QUAD,1.99 $\pm$ 0.59,\textbf{1.00} $\pm$ 0.00,0.52 $\pm$ 0.17,\textbf{0.78} $\pm$ 0.06
,FRPD-DPP-GR,2.00 $\pm$ 0.52,\textbf{1.00} $\pm$ 0.00,\textbf{0.45} $\pm$ 0.23,0.75 $\pm$ 0.17
,FRPD-DPP-LS,2.00 $\pm$ 0.52,\textbf{1.00} $\pm$ 0.00,\textbf{0.45} $\pm$ 0.23,0.75 $\pm$ 0.17
\end{filecontents*}
\pgfplotstabletypeset[
        col sep=comma,
        string type,
        every head row/.style={before row=\toprule,after row=\midrule},
        every row no 3/.style={after row=\midrule},
        every row no 7/.style={after row=\midrule},
        every row no 11/.style={after row=\midrule},
        every last row/.style={after row=\bottomrule},
        columns/dataset/.style={column name=Dataset, column type={l}},
        columns/method/.style={column name=Methods, column type={l}},
        columns/cost/.style={column name=Shortest Path, column type={c}},
        columns/valid/.style={column name=Validity, column type={c}},
        columns/diversity/.style={column name=Anti Diversity, column type={c}},
        columns/dpp/.style={column name=DPP, column type={c}},
    ]{mlp_action.csv}
\end{table*}

\textbf{Datasets.} To construct a diverse recourse plan, we rely on the premise that the available data should be sufficiently diverse in order to find a set of diverse prototypes. Thus, we pick four real-world datasets in the domain of financial applications with potential consequential decisions (e.g., credit loan): German credit~\citep{ref:dua2017uci}, Small Business Administration (SBA)~\citep{ref:li2018should}, Bank~\citep{ref:dua2017uci}, Adult~\citep{ref:dua2017uci}. The chosen datasets are widely used in the literature of diverse recourses~\citep{ref:mothilal2020explaining, ref:bui2022counterfactual, ref:haldar2021reliable, ref:wang2021skyline, ref:hasan2022data}. We preprocess the data using the same min-max standardizer for the continuous features and one-hot encoding for categorical features as in~\cite{ref:mothilal2020explaining}. The FACE-based interpolation scheme can easily handle one-hot features. For linear interpolation, we treat one-hot features in a probabilistic manner to relax them to continuous domains. The feature value with the highest probability will be activated. This approach is common in the recourse literature~\citep{ref:wachter2017counterfactual, ref:mothilal2020explaining, ref:upadhyay2021towards}.

\textbf{Classifier.} We use a three-layer MLP classifier $\mc C$ with the hidden sizes of 20, 50, and 20. For each dataset, we split uniformly at random $80\%$ of the original dataset to train the classifier~$\mc C$. The remaining data from the dataset is used to evaluate our methods and baselines.

We compare using the following metrics:

\textbf{Cost.} We measure the cost of a plan $\{x_k^{r}\}$ by the average distance from the instance $x_0$ to each recourse in the plan:
\begin{equation} \label{eq:cost}
    \nonumber \mathrm{Cost}(\{x_k^{r}\}, x_0) \Let \frac{1}{K} \sum\nolimits_{k=1}^K \mathrm{dist}(x_k^{r}, x_0).
\end{equation}
\textbf{Validity.} A plan is considered valid if every recourse in the plan flips the prediction of the underlying model. We then compute the validity metric as the fraction of instances for which the constructed recourse plan is valid.

\textbf{Anti-Diversity.} We compute the $\mathrm{Anti\text{-}Diversity}$ of a recourse plan $\{x_k^r\}$ as the sum of pairwise cosine similarity of the normalized direction vectors, which can be written
\begin{multline} \notag
    \mathrm{Anti\text{-}Diversity}(\{x_k^{r}\}, x_0)\Let 
    \\ \sum_{1 \le k,k' \le K, k \neq k'} \frac{(x_k^r - x_0)^\top (x_{k'}^r - x_0)}{\|x_k^r - x_0\|_2 \| x_{k'}^r - x_0 \|_2}.
\end{multline}
\textbf{DPP.} We compute the DPP of a recourse plan using the same measure as in~\cite{ref:mothilal2020explaining}, which can be written explicitly as
\[
    \begin{array}{cll}
         & \mathrm{DPP}(\{x_k^r\}) \Let \det(Q),  \\ \textrm{where} \quad & Q_{ij} = (1 + \mathrm{dist}(x_i^r, x_j^r))^{-1} \quad \forall 1 \leq i, j \leq K.
    \end{array}
\]
\textbf{Distance to data manifold.} The distance of a recourse plan to the data manifold is the maximum distance from each recourse in the plan to the data manifold. Notice that measuring the adherence to the data manifold using a latent distance is common in disentangled-based recourse methods such as~\citet{ref:pawelczyk2020learning}. In this experiment, we assume data from class $1$ is the target data manifold:
\begin{equation} \notag
    \begin{array}{cll}
         & \mathrm{Manifold\text{-}Distance}(\{x_k^{r}\}) \Let \\
         &  \max_{k \in \{1, \ldots, K\}} \min_{i \in \{1 , \ldots, N\}}~\mathrm{dist}(x_k^{r}, \wh x_i).
    \end{array}
    \label{eq:manifold_dist}
\end{equation}


\subsection{Experiments without Actionability Constraints}
To evaluate the trade-off between cost and Anti-Diversity, and between cost and DPP, we vary $\theta = \vartheta \in [0.1, 1.0]$ for two methods: FRPD-QUAD, FRPD-DPP-LS. For DiCE's parameters, we fix the proximity weight and change the diversity weight between $0$ and $5.0$.  Because there are no actionability constraints, we choose the distance function as the Euclidean distance. 

We visualize the Pareto frontiers of the cost-Anti-Diversity and cost-DPP trade-offs of FRPD-QUAD, FRPD-DPP-LS, and DiCE in Figure~\ref{fig:mlp_pareto}. We report the additional results for FRPD-DPP-GR in Appendix~\ref{sec:app-exp}. The results in Figure~\ref{fig:mlp_pareto} demonstrate that FRPD-QUAD, FRPD-DPP-GR, and FRPD-DPP-LS produce a diverse recourse plan at a far lower cost than DiCE. In addition, Table~\ref{tab:manifold_dist} demonstrates that the validity of DiCE drops to below one in the SBA dataset. Further, FRPD-QUAD has the lowest Anti-Diversity, which is natural because this method optimizes the Anti-Diversity metric explicitly. Similarly, if we use the DPP to measure the diversity, our DPP-based methods, such as FRPD-DPP-GR and FRPD-DPP-LS outperform in this metric for three over four real datasets. Finally, our three methods are more likely to generate a recourse plan that is close to the data manifold.

\subsection{Experiments with Actionability Constraints}
Herein, we use the shortest path interpolation on the actionability graph and compare our recourses against FACE. We use CARLA's source code~\citep{ref:pawelczyk2021carla} to construct the actionability graph as the common input.

In this benchmark, we choose the distance function as the shortest path on the actionability graph.
Because each recourse has to be a node in the graph, we do not compare the distance to the manifold in this experiment. The results in Table~\ref{tab:action} demonstrate that our three methods outperform FACE in terms of two diversity metrics, Anti-Diversity and DPP. This result is natural because our methods directly optimize for the Anti-Diversity metric. In contrast, FACE does not take diversity as an explicit criterion. The FACE method, however, provides the smallest cost.

\begin{figure*}[!ht]
    \centering
    \includegraphics[width=0.6\linewidth]{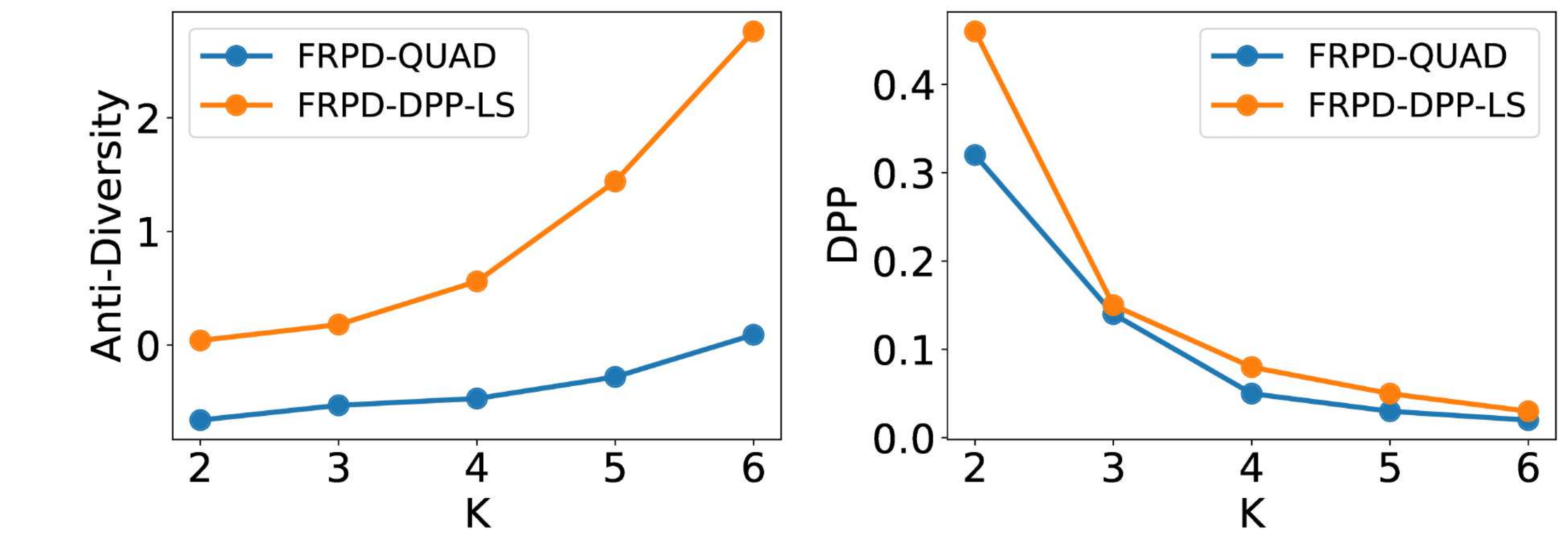}
    \caption{Impact of the number of recourses $K$ to Anti-Diversity and DPP on the German Credit dataset.}
    \label{fig:impact_k}
    \vspace{-5mm}
\end{figure*}

\subsection{Sensitivity analysis of the number of recourses $K$} In this experiment, we study the impact of the number of selected prototypes $K$ of a recourse plan on two diversity metrics: Anti-Diversity and DPP. First, we choose $\theta = \vartheta = 0.9$ and vary the number of recourses in a plan $K \in [2, 6]$ for our two methods: FRPD-QUAD and FRPD-DPP-LS.

Figure~\ref{fig:impact_k} reports the trade-off between $K$ and two diversity measures on the German credit dataset. These results show that there is a trade-off between the number of selected prototypes and the diversity measures: as $K$ increases, Anti-Diversity increases and DPP decreases.

\textbf{Conclusions.} We propose a novel method to generate a diverse recourse plan that is feasible in the sense that the plan takes into consideration multiple criteria such as cost, validity, diversity, and adherence to the data manifold. As diversity may possess different meanings and metrics, we provide two independent and possibly complementary approaches to select the prototypes for subsequent interpolation: one relies on MAP inference for DPPs and the other relies on a quadratic programming formulation. The prototypes can be used as targets for different interpolation schemes to generate diverse and sequential recourses. Finally, we remark that the goal of this paper is \textit{not} about providing a normative answer to which diversity metric should be preferred. We instead relegate this interesting question to future research. 

\begin{remark}[Privacy of recourse] \label{rem:extension}
    The goal of this paper is to devise a diverse recourse plan that has a higher capacity of matching the different preferences of the subjects.
    
    A foundational assumption of this paper is the available access to the training data. In particular, to construct the set of diverse prototypes, we need access to positively-predicted training samples. The quality of our approach is dependent on the instances with positive predictions. Given that recourse is user-dependent, the training set for the problem of recourse diversity should be sufficiently diverse to account for a wide range of user preferences. 
    
    Moreover, if we use the graph interpolation in Section~\ref{sec:network}, then we need additional access to the negatively-predicted training samples. Access to the training data is crucial to identify the manifold of the data and to generate actionable recourses. This access to the training data is leveraged in existing methods such as FACE~\citep{ref:poyiadzi2020face}. The required access to the training data is often criticized due to its potential breach of \textit{privacy}. To resolve this issue, we can utilize multiple techniques and methods to guarantee privacy. For a concrete example, each sample can be modified by adding an independent noise vector, which is able to satisfy the notion of local differential privacy. The exact computation of the noise variance is left for future research.
\end{remark}

\begin{remark}[Family of classifiers]
    In this work, we treat the underlying classifier as a black-box model, so our methods can handle any family of classifiers (deep NNs, tree-based classifiers, etc.). Moreover, our framework can easily adapt to the multi-class classification problem. In this case, we treat the data from the desired class as potential candidates for selecting $K$ prototypes. An interpolation technique between the input instance and $K$ prototypes can be used to find recourses, for example, by linear interpolation.
\end{remark}

\textbf{Acknowledgments.} Viet Anh Nguyen acknowledges the generous support from the CUHK's Improvement on Competitiveness in Hiring New Faculties Funding Scheme.


\newpage
\onecolumn




\appendix
\onecolumn
\aistatstitle{Supplementary Materials for Paper: Feasible Recourse Plan via Diverse Interpolation}

\section{Additional Experiment Results} \label{sec:app-exp}

Source code, datasets, and results can be accessed from \url{https://github.com/duykhuongnguyen/recourse-plan-diverse-interpolation}. We use the implementation of CARLA~\citep{ref:pawelczyk2021carla} for two baselines: DiCE and FACE.

\subsection{Datasets}

\textbf{Real-world datasets.} To form a diverse recourse plan, we rely on the premise that the available data should be sufficiently diverse in order to find a set of diverse prototypes. Thus, we pick four real-world datasets in the domain of financial applications with potential consequential decisions (e.g., credit loan) and these datasets are diverse in different perspectives:

\begin{itemize}[leftmargin = 5mm]
    \item The German Credit dataset~\citep{ref:dua2017uci} contains 1,000 samples of loan applicants, each applicant is classified as either a good or a bad client. We choose five features: Status, Duration, Credit amount, Personal status, and Age. The Duration and Credit amount features represent key attributes of a loan application or, equivalently, of a financial product. This dataset exhibits product diversity.
    \item The Small Business Administration (SBA) dataset~\citep{ref:li2018should} contains 2,102 samples of small company loans in California, the label indicates whether a business has defaulted on a loan. For the SBA dataset, we choose the following features: Selected, Term, NoEmp, CreateJob, RetainedJob, UrbanRural, ChgOffPrinGr, GrAppv, SBA Appv, New, RealEstate, Portion, Recession. Collected over 24 years from 1989 and 2012, this dataset exhibits temporal diversity.
    \item The Bank dataset~\citep{ref:dua2017uci} contains 4,522 samples of individuals undergoing direct marketing activities from a financial institution. The label represents whether the client subscribes to a bank term deposit (a financial product). We use the following features: Age, Education, Balance, Housing, Loan, Campaign, Previous, and Outcome. The Campaign and Previous feature the number of contacts with the client in this and the previous campaign, respectively, and they represent the aggressiveness in the marketing strategy. This dataset exhibits strategic diversity.
    \item The Adult dataset~\citep{ref:dua2017uci} contains 32,560 samples of individuals for income prediction. For this dataset, we use full features of the dataset: Age, Workclass, Fnlwgt, Education, Educational-num, Marital-status, Occupation, Relationship, Race, Gender, Capital-gain, Capital-loss, Hours-per-week, Native-country, and Income. This dataset collects a diverse set of individual-level features, and it exhibits intersectional diversity.
\end{itemize}
The chosen datasets are widely used in the literature of diverse recourses~\citep{ref:mothilal2020explaining, ref:bui2022counterfactual, ref:haldar2021reliable, ref:wang2021skyline, ref:hasan2022data}. We preprocess the data using the same min-max standardizer for the continuous features and one-hot encoding for categorical features as in~\citet{ref:mothilal2020explaining}. 

To construct the graph for actionability methods, we choose the below set of immutable features for each dataset:

\begin{itemize}[leftmargin=5mm]
        \item For German Credit, we select ``Personal status" since it is challenging to compel changes in an individual's personal status (male, female, single, married)~\citep{ref:ustun2019actionable, ref:karimi2020algorithmic}.
        \item For SBA, we select ``Recession" and ``UrbanRural" as they are problematic to change in the foreseeable future for the SBA dataset.
        \item For Bank, we select ``previous", ``campaign", and ``outcome" since they are historical attributes.
        \item For Adult, we select ``marital status" and ``gender".
\end{itemize}

\subsection{Experimental setup}
\textbf{Classifier.} For each dataset, we first do the 80-20 split (80\% for training, 20\% for testing) and train an MLP classifier on the training set. We report the Accuracy and AUC of the MLP classifier on each dataset in Table~\ref{tab:acc_clf}.

\begin{table*}[htb]
    \centering
    \caption{Accuracy and AUC of the MLP classifiers on four real-world datasets.}
    \label{tab:acc_clf}
    \footnotesize
    \begin{filecontents*}{accuracy.csv}
data,acc,auc
German Credit,0.73,0.71
SBA,0.97,0.99
Bank,0.89,0.68
Adult,0.83,0.88
\end{filecontents*}
\pgfplotstabletypeset[
        col sep=comma,
        string type,
        every head row/.style={before row=\toprule,after row=\midrule},
        every last row/.style={after row=\bottomrule},
        columns/data/.style={column name=Dataset, column type={l}},
        columns/acc/.style={column name=Accuracy, column type={l}},
        columns/auc/.style={column name=AUC, column type={c}},
    ]{accuracy.csv}
\end{table*}

\textbf{Settings for Figure~\ref{fig:mlp_pareto}.} In this experiment, we generate a recourse plan with $K=3$ recourses. We vary $\theta = \vartheta \in [0.1, 1.0]$ for three methods: FRPD-QUAD, FRPD-DPP-GR and FRPD-DPP-LS. We choose $h=1.0$ for FRPD-DPP-GR and FRPD-DPP-LS. We fix the proximity weight $0.5$ as the default setting and change the diversity weight between $0$ and $5.0$ for DiCE.

\textbf{Settings for Table~\ref{tab:manifold_dist} and Table~\ref{tab:action}.} In this experiment, we choose $K=3$ and $\theta = \vartheta = 0.9$ for our three methods: FRPD-QUAD, FRPD-DPP-GR and FRPD-DPP-LS. We choose $h=1.0$ for FRPD-DPP-GR and FRPD-DPP-LS. We use the default setting for proximity weight and diversity weight of DiCE with $0.5$ and $1.0$, respectively. We use the default settings in CARLA~\citep{ref:pawelczyk2021carla} for FACE.

\subsection{Additional numerical results}

\begin{figure*}[!ht]
    \centering
    \includegraphics[width=\linewidth]{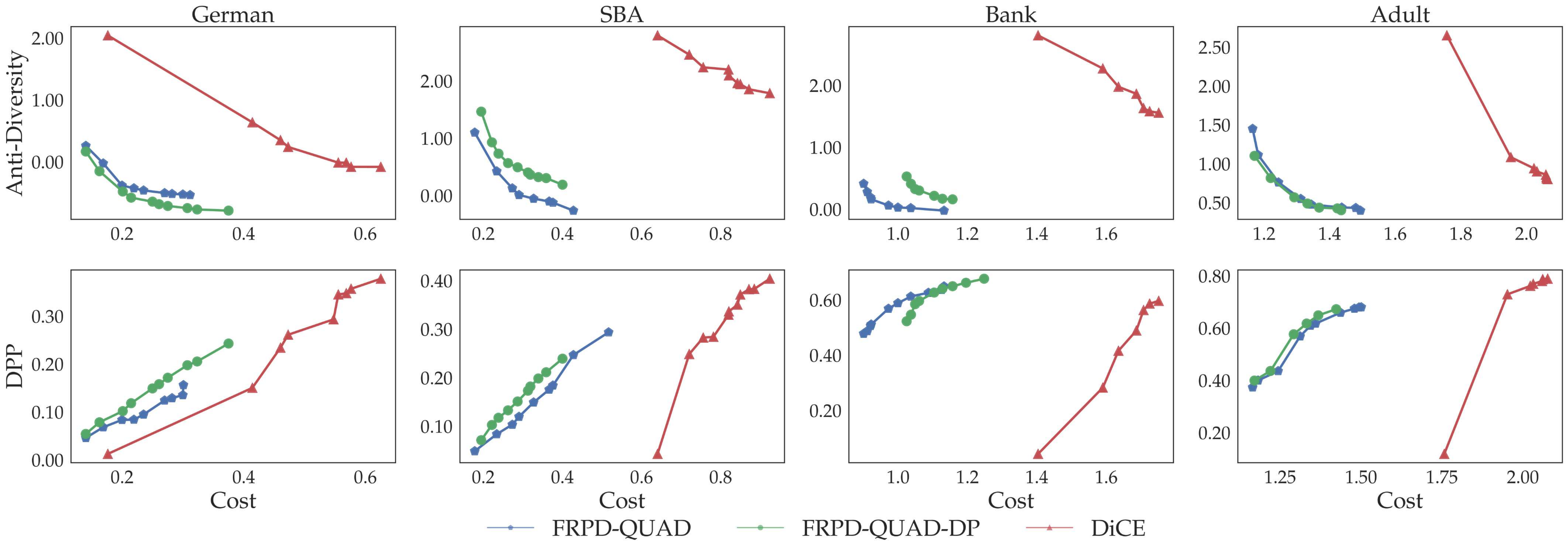}
    \caption{Pareto frontier of the cost vs.~Anti-Diversity (top) and cost vs.~DPP (bottom) trade-off between FRPD-QUAD, FRPD-QUAD-DP using linear interpolation, and DiCE on four real-world datasets.}
    \label{fig:mlp_pareto_dp}
\end{figure*}

\textbf{Dual program iterations.}
In the main paper, we utilize the best response iterations in Algorithm~\ref{alg:br} (denoted FRPD-QUAD) to solve problem~\eqref{eq:minmax}. In this section, we use the dual program iterations in Algorithm~\ref{alg:dual} (denoted FRPD-QUAD-DP) to solve the same problem and compare it with FRPD-QUAD. We evaluate different trade-offs: cost vs.~Anti-Diversity and cost vs.~DPP trade-off on four real-world datasets, and report the results in Figure~\ref{fig:mlp_pareto_dp}. We also compute cost, Anti-Diversity, DPP, manifold distance, and discuss how each metric is correlated, and report the results in Table~\ref{tab:manifold_dist_br_dp} and Table~\ref{tab:action_dp}. The settings for these experiments are the same as in Figure~\ref{fig:mlp_pareto}, Table~\ref{tab:manifold_dist}, and Table~\ref{tab:action}.

The results in Figure~\ref{fig:mlp_pareto_dp} indicate that FRPD-QUAD-DP has higher DPP than FRPD-QUAD in three datasets. Further, the recourse plans generated by FRPD-QUAD are more directly correlated with the data manifold in the SBA and Bank datasets. Our two approaches also produce a recourse plan that closely resembles the data manifold. Table~\ref{tab:action_dp} shows that in all four datasets, FRPD-QUAD-DP outperforms FACE in terms of Anti-Diversity and DPP. This implies that FRPD-QUAD-DP recourses are more diverse than FACE recourses.

\begin{table*}[htb]
    \centering
    \caption{Benchmark of Cost, Validity, Anti-Diversity, DPP and Manifold-Distance between FRPD-QUAD, FRPD-DP using linear interpolation, and DiCE on four real-world datasets.}
    \label{tab:manifold_dist_br_dp}
    \footnotesize
    \begin{filecontents*}{mlp_br_dp.csv}
dataset,method,cost,valid,diversity,dpp,manifold_dist
German,DiCE,0.41 $\pm$ 0.34,\textbf{1.00} $\pm$ 0.00,0.71 $\pm$ 0.74,0.14 $\pm$ 0.09,\textbf{0.15} $\pm$ 0.11
,FRPD-QUAD,\textbf{0.30} $\pm$ 0.19,\textbf{1.00} $\pm$ 0.00,-0.53 $\pm$ 0.70,0.14 $\pm$ 0.12,0.19 $\pm$ 0.19
,FRPD-QUAD-DP,0.31 $\pm$ 0.23,\textbf{1.00} $\pm$ 0.00,\textbf{-0.75} $\pm$ 0.61,\textbf{0.20} $\pm$ 0.15,0.17 $\pm$ 0.18
SBA,DiCE,0.71 $\pm$ 0.22,0.95 $\pm$ 0.22,2.49 $\pm$ 0.61,0.23 $\pm$ 0.12,0.86 $\pm$ 0.24
,FRPD-QUAD,0.45 $\pm$ 0.30,\textbf{1.00} $\pm$ 0.00,\textbf{-0.17} $\pm$ 0.60,\textbf{0.23} $\pm$ 0.19,\textbf{0.23} $\pm$ 0.16
,FRPD-QUAD-DP,\textbf{0.34} $\pm$ 0.28,\textbf{1.00} $\pm$ 0.00,0.32 $\pm$ 0.65,0.20 $\pm$ 0.18,0.35 $\pm$ 0.25
Bank,DiCE,1.62 $\pm$ 0.32,\textbf{1.00} $\pm$ 0.00,2.14 $\pm$ 0.64,0.34 $\pm$ 0.26,\textbf{0.48} $\pm$ 0.37
,FRPD-QUAD,\textbf{1.09} $\pm$ 0.21,\textbf{1.00} $\pm$ 0.00,\textbf{0.03} $\pm$ 0.26,0.63 $\pm$ 0.11,0.51 $\pm$ 0.20
,FRPD-QUAD-DP,1.20 $\pm$ 0.25,\textbf{1.00} $\pm$ 0.00,0.16 $\pm$ 0.31,\textbf{0.66} $\pm$ 0.09,0.60 $\pm$ 0.22
Adult,DiCE,1.95 $\pm$ 0.37,\textbf{1.00} $\pm$ 0.00,1.08 $\pm$ 0.55,\textbf{0.73} $\pm$ 0.13,1.54 $\pm$ 0.31
,FRPD-QUAD,1.48 $\pm$ 0.58,\textbf{1.00} $\pm$ 0.00,0.43 $\pm$ 0.33,0.67 $\pm$ 0.18,0.60 $\pm$ 0.32
,FRPD-QUAD-DP,\textbf{1.44} $\pm$ 0.59,\textbf{1.00} $\pm$ 0.00,\textbf{0.40} $\pm$ 0.36,0.67 $\pm$ 0.18,\textbf{0.59} $\pm$ 0.31
\end{filecontents*}
\pgfplotstabletypeset[
        col sep=comma,
        string type,
        every head row/.style={before row=\toprule,after row=\midrule},
        every row no 2/.style={after row=\midrule},
        every row no 5/.style={after row=\midrule},
         every row no 8/.style={after row=\midrule},
        every last row/.style={after row=\bottomrule},
        columns/dataset/.style={column name=Dataset, column type={l}},
        columns/method/.style={column name=Methods, column type={l}},
        columns/cost/.style={column name=Cost, column type={c}},
        columns/valid/.style={column name=Validity, column type={c}},
        columns/diversity/.style={column name=Anti Diversity, column type={c}},
        columns/dpp/.style={column name=DPP, column type={c}},
        columns/manifold_dist/.style={column name=Manifold Distance, column type={c}},
    ]{mlp_br_dp.csv}
\end{table*}

\begin{table*}[htb]
    \centering
    \caption{Benchmark of Cost, Validity, Anti-Diversity, and DPP between FRPD-QUAD, FRPD-QUAD-DP, and FACE with actionability constraints on four real-world datasets.}
    \label{tab:action_dp}
    \footnotesize
    \begin{filecontents*}{mlp_action_frpd_dpp.csv}
dataset,method,cost,valid,diversity,dpp
German,FACE,\textbf{0.24} $\pm$ 0.17,\textbf{1.00} $\pm$ 0.00,2.18 $\pm$ 0.77,0.04 $\pm$ 0.04
,FRPD-QUAD,0.55 $\pm$ 0.24,\textbf{1.00} $\pm$ 0.00,\textbf{-0.13} $\pm$ 0.24,0.21 $\pm$ 0.07
,FRPD-QUAD-DP,0.33 $\pm$ 0.11,\textbf{1.00} $\pm$ 0.00,0.20 $\pm$ 0.23,\textbf{0.24} $\pm$ 0.05
SBA,FACE,\textbf{0.25} $\pm$ 0.20,\textbf{1.00} $\pm$ 0.00,2.08 $\pm$ 0.78,0.02 $\pm$ 0.03
,FRPD-QUAD,0.55 $\pm$ 0.25,\textbf{1.00} $\pm$ 0.00,\textbf{-0.04} $\pm$ 0.41,0.17 $\pm$ 0.09
,FRPD-QUAD-DP,0.68 $\pm$ 0.32,\textbf{1.00} $\pm$ 0.00,0.35 $\pm$ 0.25,\textbf{0.21} $\pm$ 0.09
Bank,FACE,\textbf{1.29} $\pm$ 0.29,\textbf{1.00} $\pm$ 0.00,1.42 $\pm$ 0.86,0.31 $\pm$ 0.27
,FRPD-QUAD,1.55 $\pm$ 0.44,\textbf{1.00} $\pm$ 0.00,\textbf{0.06} $\pm$ 0.12,0.69 $\pm$ 0.16
,FRPD-QUAD-DP,1.48 $\pm$ 0.55,\textbf{1.00} $\pm$ 0.00,0.07 $\pm$ 0.16,\textbf{0.72} $\pm$ 0.03
Adult,FACE,\textbf{1.33} $\pm$ 0.36,\textbf{1.00} $\pm$ 0.00,2.66 $\pm$ 0.36,0.22 $\pm$ 0.27
,FRPD-QUAD,1.99 $\pm$ 0.59,\textbf{1.00} $\pm$ 0.00,\textbf{0.52} $\pm$ 0.17,\textbf{0.78} $\pm$ 0.06
,FRPD-QUAD-DP,1.98 $\pm$ 0.52,\textbf{1.00} $\pm$ 0.00,0.56 $\pm$ 0.19,0.75 $\pm$ 0.04

\end{filecontents*}
\pgfplotstabletypeset[
        col sep=comma,
        string type,
        every head row/.style={before row=\toprule,after row=\midrule},
        every row no 2/.style={after row=\midrule},
        every row no 5/.style={after row=\midrule},
        every row no 8/.style={after row=\midrule},
        every last row/.style={after row=\bottomrule},
        columns/dataset/.style={column name=Dataset, column type={l}},
        columns/method/.style={column name=Methods, column type={l}},
        columns/cost/.style={column name=Shortest Path, column type={c}},
        columns/valid/.style={column name=Validity, column type={c}},
        columns/diversity/.style={column name=Anti Diversity, column type={c}},
        columns/dpp/.style={column name=DPP, column type={c}},
    ]{mlp_action_frpd_dpp.csv}
\end{table*}

\begin{figure*}[!ht]
    \centering
    \includegraphics[width=\linewidth]{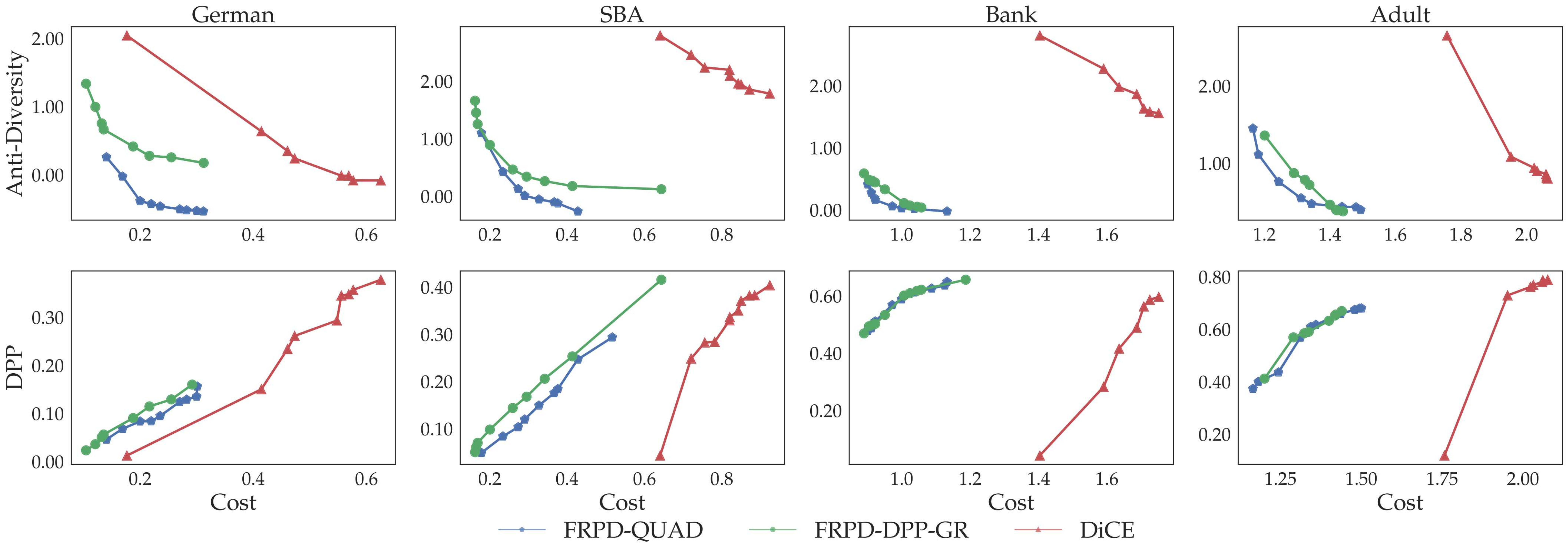}
    \caption{Pareto frontier of the cost vs.~Anti-Diversity (top) and cost vs.~DPP (bottom) trade-off between FRPD-QUAD, FRPD-DPP-GR using linear interpolation, and DiCE on four real-world datasets.}
    \label{fig:mlp_pareto_dppgr}
\end{figure*}

\textbf{Cost-diversity trade-off of FRPD-DPP-GR.} Here, we provide the additional results for FRPD-DPP-GR: the cost and Anti-Diversity, cost, and DPP trade-off of FRPD-DPP-GR without actionability constraints. We use the same settings for FRPD-DPP-GR: fix $h = 1.0$ and vary $\vartheta \in [0.1, 1.0]$ . The results in Figure~\ref{fig:mlp_pareto_dppgr} indicate that FRPD-QUAD and FRPD-DPP-GR produce a diverse recourse plan at a far lower cost than DiCE.

\textbf{Results on more baselines and COMPAS dataset.} We provide additional experiments with a new dataset: Compas. We also conduct the experiments with three additional baselines: LORE~\citep{ref:guidotti2018local} and CCHVAE~\citep{ref:pawelczyk2020learning} for non-actionability comparison and CRUDS~\citep{ref:downs2020cruds} for actionability comparison. We report the results for the German and Compas datasets in Table~\ref{tab:reb_non} and Table~\ref{tab:reb_action}. Our method outperforms LORE, CCHVAE, and CRUDS in terms of Anti-Diversity and DPP.

\begin{table*}[htb]
    \centering
   \caption{Comparing our method (QUAD) versus LORE and CCHVAE on German and Compas data with\textit{out} actionability constraint.}
    \label{tab:reb_non}
    \footnotesize
   \begin{filecontents*}{mlp_rebuttal_non.csv}
dataset,method,cost,valid,diversity,dpp,manifold_dist
German,LORE,\textbf{0.19} $\pm$ 0.05,0.87 $\pm$ 0.27,1.05 $\pm$ 0.74,0.03 $\pm$ 0.02,\textbf{0.12} $\pm$ 0.10
,CCHVAE,0.49 $\pm$ 0.12,\textbf{1.00} $\pm$ 0.00,0.95 $\pm$ 0.42,0.05 $\pm$ 0.02,0.25 $\pm$ 0.18
,FRPD-QUAD,0.30 $\pm$ 0.19,\textbf{1.00} $\pm$ 0.00,\textbf{-0.53} $\pm$ 0.70,\textbf{0.14} $\pm$ 0.12,0.19 $\pm$ 0.19
Compas,LORE,\textbf{0.31} $\pm$ 0.22,0.93 $\pm$ 0.19,1.13 $\pm$ 0.91,0.03 $\pm$ 0.05,0.08 $\pm$ 0.11
,CCHVAE,0.55 $\pm$ 0.45,\textbf{1.00} $\pm$ 0.00,1.45 $\pm$ 1.05,0.07 $\pm$ 0.06,0.13 $\pm$ 0.17
,FRPD-QUAD,0.37 $\pm$ 0.25,\textbf{1.00} $\pm$ 0.00,\textbf{-0.71} $\pm$ 0.31,\textbf{0.24} $\pm$ 0.20,\textbf{0.07} $\pm$ 0.09
\end{filecontents*}
\pgfplotstabletypeset[
        col sep=comma,
        string type,
        every head row/.style={before row=\toprule,after row=\midrule},
        every row no 2/.style={after row=\midrule},
        every row no 5/.style={after row=\midrule},
        every row no 11/.style={after row=\midrule},
        every last row/.style={after row=\bottomrule},
        columns/dataset/.style={column name=Dataset, column type={l}},
        columns/method/.style={column name=Methods, column type={l}},
        columns/cost/.style={column name=Cost, column type={c}},
        columns/valid/.style={column name=Validity, column type={c}},
        columns/diversity/.style={column name=Anti Diversity, column type={c}},
        columns/dpp/.style={column name=DPP, column type={c}},
        columns/manifold_dist/.style={column name=Manifold Distance, column type={c}},
    ]{mlp_rebuttal_non.csv}
\end{table*}

\begin{table*}[htb]
    \centering
    \caption{Comparing our method (QUAD) versus FACE and CRUDS on two datasets with actionability constraint}
    \label{tab:reb_action}
    \footnotesize
   \begin{filecontents*}{mlp_rebuttal_action.csv}
dataset,method,cost,valid,diversity,dpp
German,FACE,\textbf{0.24} $\pm$ 0.17,\textbf{1.00} $\pm$ 0.00,2.18 $\pm$ 0.77,0.04 $\pm$ 0.04
,CRUDS,0.78 $\pm$ 0.32,\textbf{1.00} $\pm$ 0.00,1.56 $\pm$ 0.42,0.11 $\pm$ 0.07
,FRPD-QUAD,0.55 $\pm$ 0.24,\textbf{1.00} $\pm$ 0.00,\textbf{-0.13} $\pm$ 0.24,\textbf{0.21} $\pm$ 0.07
Compas,FACE,\textbf{0.35} $\pm$ 0.26,\textbf{1.00} $\pm$ 0.00,2.23 $\pm$ 0.83,0.07 $\pm$ 0.05
,CRUDS,0.89 $\pm$ 0.47,\textbf{1.00} $\pm$ 0.00,1.92 $\pm$ 0.56,0.14 $\pm$ 0.09
,FRPD-QUAD,0.59 $\pm$ 0.34,\textbf{1.00} $\pm$ 0.00,\textbf{-0.25} $\pm$ 0.17,\textbf{0.32} $\pm$ 0.12
\end{filecontents*}
\pgfplotstabletypeset[
        col sep=comma,
        string type,
        every head row/.style={before row=\toprule,after row=\midrule},
        every row no 2/.style={after row=\midrule},
        every row no 5/.style={after row=\midrule},
        every row no 11/.style={after row=\midrule},
        every last row/.style={after row=\bottomrule},
        columns/dataset/.style={column name=Dataset, column type={l}},
        columns/method/.style={column name=Methods, column type={l}},
        columns/cost/.style={column name=Shortest Path, column type={c}},
        columns/valid/.style={column name=Validity, column type={c}},
        columns/diversity/.style={column name=Anti Diversity, column type={c}},
        columns/dpp/.style={column name=DPP, column type={c}},
    ]{mlp_rebuttal_action.csv}
\end{table*}

\textbf{A recourse plan example.} We present an example of a recourse plan using the German credit dataset with $K=3$ recourses between two methods without actionability constraints (DiCE and FRPD-QUAD using linear interpolation). Two approaches (FRPD-QUAD and DiCE) might present users with a diverse set of recourses to choose from. However, we can observe in Table~\ref{tab:recourse_exp} that recourse plans generated by our method are more realistic and suited for users to make changes. For example, DiCE suggests a recourse that involves changing the age from 22 to 20.1 years old, which is not a feasible change.

\begin{table*}[htb]
    \centering
    \caption{An example of a recourse plan generated by FRPD-QUAD using linear interpolation and DiCE on the German Credit dataset.}
    \label{tab:recourse_exp}
    \footnotesize
    \begin{filecontents*}{recourse_examples.csv}
method,Duration,Credit Amount,Personal Status,Age
Input,12,1007,A94,22
FRPD-QUAD,24.2,2918.9,A94,22
,10.9,943.9,A94,22.2
,10.7,682.1,A94,22
DiCE,11.9,1304.4,A91,20.1
,4,250,A94,22
,15.4,1073.5,A94,32.5
\end{filecontents*}
\pgfplotstabletypeset[
        col sep=comma,
        string type,
        every head row/.style={before row=\toprule,after row=\midrule},
        every row no 0/.style={after row=\midrule},
        every row no 3/.style={after row=\midrule},
        every last row/.style={after row=\bottomrule},
        columns/method/.style={column name=, column type={l}},
    ]{recourse_examples.csv}
\end{table*}

\clearpage
\subsection{Sequential Recourse}

\begin{figure*}[!ht]
    \centering
    \includegraphics[width=0.8\linewidth]{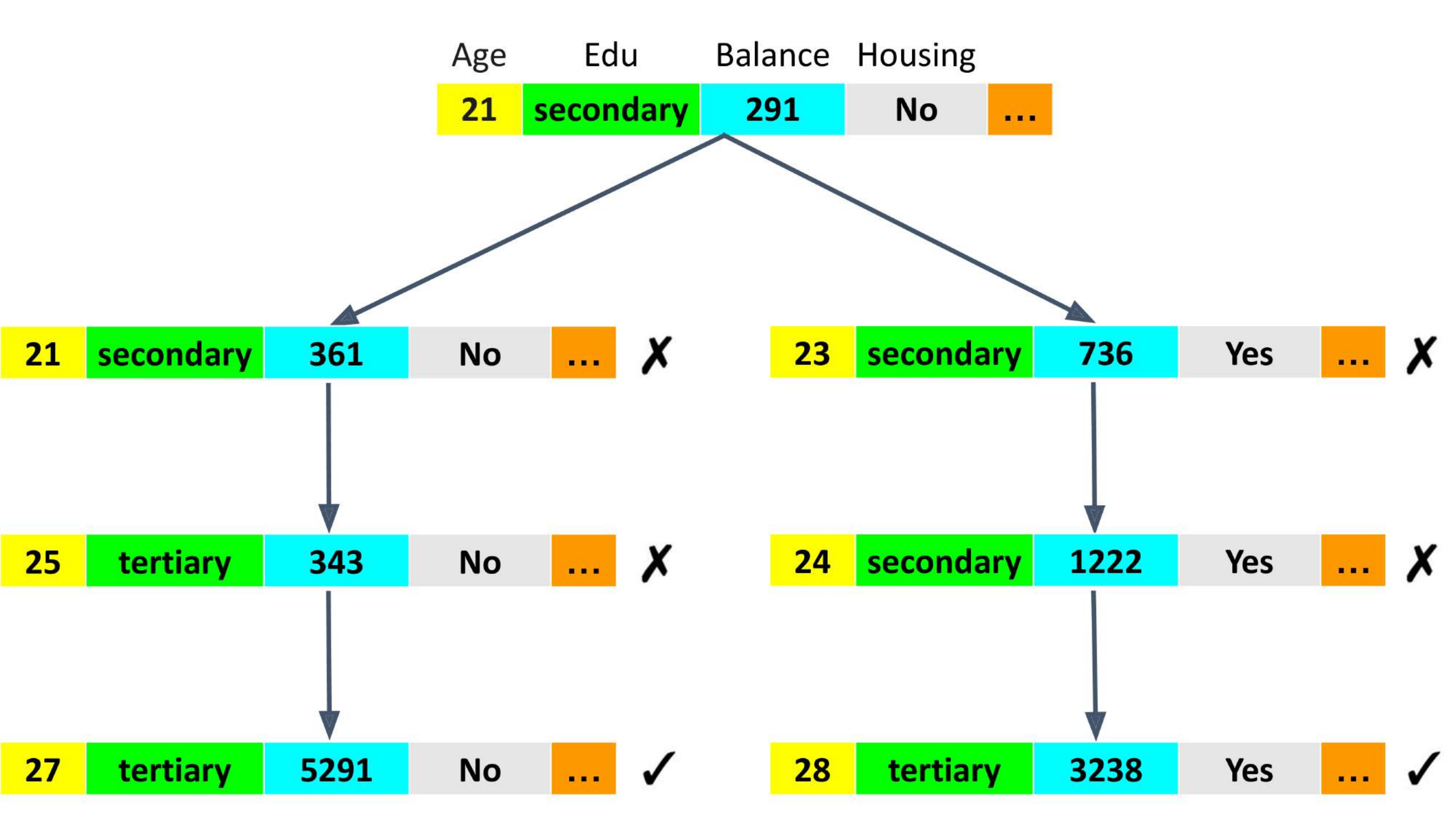}
    \caption{An example of a sequential recourse plan with $K=2$ recourses on the Bank dataset. \xmark \space and \cmark \space denote the unfavorable and favorable predicted  outcomes, respectively.}
    \label{fig:recourse_exp_seq}    
\end{figure*}

The original recourse with single shot recommendations~\citep{ref:russell2019efficient, ref:mothilal2020explaining} provide `what-if' feedback for users. However, in practice, it is more desirable to provide a sequence of directive actions that users should take in order to achieve that recourse~\citep{ref:singh2021directive, ref:verma2022amortized}. Sequential actions are also more realistic than a one-step change in real-world applications~\citep{ref:ramakrishnan2020synthesizing}. We present an example of a sequential recourse plan with $K=2$ recourses on the Bank dataset in Figure~\ref{fig:recourse_exp_seq}.

\textbf{Diversity of Sequential Recourses.} Similar to the original recourse plan~\citep{ref:mothilal2020explaining}, it is also desirable to promote diversity among suggested sequential recourses in order to capture multiple users' preferences. To evaluate the diversity of sequential recourse plans, we measure $\mathrm{Path\text{-}Diveristy}$ using Levenshtein distance~\citep{ref:navarro2001guided} and $\mathrm{Path\text{-}Anti\text{-}Diversity}$ using Jaccard coefficient~\citep{ref:chondrogiannis2018finding}.

 
\begin{itemize}
    \item \textbf{Path Diversity}: The diversity of a sequential recourse plan is computed by
    \[
        \mathrm{Path\text{-}Diversity}(P_1, \ldots, P_k) = \frac{1}{{K \choose 2}} \sum_{\substack{i, j \in [K] \\ i < j}} \mathrm{lev} (P_i, P_j).
    \]
    Here, $\mathrm{lev}(\cdot, \cdot)$ is the Levenshtein distance between two paths defined by sequences of nodes $P = (u_0, \ldots, u_l)$ and $Q = (v_0, \ldots, v_h)$. The Levenshtein distance is given by
    \[
        \mathrm{Lev} ( P, Q ) = \min
        \left\{\begin{array}{cl}
            & d(v_h, v_{h-1}) + \mathrm{Lev}(P, Q \setminus v_h), \\
            & d(u_l, u_{l-1}) + \mathrm{Lev}(P \setminus u_l, Q),  \\
            & d(u_l, v_h) + \mathrm{Lev}(P \setminus u_l, Q \setminus v_h)
        \end{array} \right\},
    \] 
    where $u_i, v_j$ are nodes on the graph $\mc G$ and the function $d(\cdot, \cdot)$ is the Euclidean distance between node features. We set $\mathrm{lev}(\emptyset, \emptyset) = 0$.
    \item \textbf{Path Anti Diversity}: The anti-diversity of a sequential recourse plan is the average Jaccard coefficient of all pairs of sequential recourses:
    \[
        \mathrm{Path\text{-}Anti\text{-}Diversity}(P_1, \ldots, P_k) = \frac{1}{{K \choose 2}} \sum_{\substack{i, j \in [K] \\ i < j}} \mathrm{Jac} (P_i, P_j).
    \]
    Here, the Jaccard coefficient defined on edges of two paths $P = (e_0, \ldots, e_l)$ and $Q = (o_0, \ldots, o_h)$ is given by
    \[
        \mathrm{Jac} (P, Q) = \frac{\sum_{e \in P \cap Q} d(e)}{\sum_{e \in P \cup Q} d(e)},
    \]
    where $d(e)$ is the Euclidean distance between two vertices of the edge $e$.
\end{itemize}


Table~\ref{tab:action_diverse_path} reports the Path-Diversity, Path-Anti-Diversity, and cost of recourse by the shortest paths connecting the input instance to the prototypes. It can be seen that sequential recourse plans generated by our methods (FRPD-QUAD, FRPD-DPP-GR, and FRPD-DPP-LS) are more diverse than those generated by FACE. 

\begin{table*}[htb]
    \centering
    \caption{Benchmark of Path-Diversity and Path-Anti-Diversity between competing methods with actionability constraints on four real-world datasets.}
    \label{tab:action_diverse_path}
    \footnotesize
    \begin{filecontents*}{mlp_diverse_path_updated.csv}
dataset,method,cost,lev,jac
German,FACE,\textbf{0.24} $\pm$ 0.17,1.07 $\pm$ 0.20,0.06 $\pm$ 0.12
,FRPD-QUAD,0.55 $\pm$ 0.24,\textbf{2.00} $\pm$ 0.67,0.05 $\pm$ 0.07
,FRPD-DPP-GR,0.42 $\pm$ 0.20,1.77 $\pm$ 0.72,\textbf{0.02} $\pm$ 0.06
,FRPD-DPP-LS,0.42 $\pm$ 0.20,1.77 $\pm$ 0.72,\textbf{0.02} $\pm$ 0.06
SBA,FACE,\textbf{0.25} $\pm$ 0.20,1.90 $\pm$ 0.70,0.06 $\pm$ 0.08
,FRPD-QUAD,0.55 $\pm$ 0.25,\textbf{5.03} $\pm$ 1.43,\textbf{0.00} $\pm$ 0.00
,FRPD-DPP-GR,0.52 $\pm$ 0.25,4.47 $\pm$ 1.54,\textbf{0.00} $\pm$ 0.00
,FRPD-DPP-LS,0.52 $\pm$ 0.25,4.47 $\pm$ 1.54,\textbf{0.00} $\pm$ 0.00
Bank,FACE,\textbf{1.29} $\pm$ 0.29,2.47 $\pm$ 0.99,0.04 $\pm$ 0.07
,FRPD-QUAD,1.55 $\pm$ 0.44,4.17 $\pm$ 0.92,0.01 $\pm$ 0.02
,FRPD-DPP-GR,1.52 $\pm$ 0.42,\textbf{4.50} $\pm$ 1.38,\textbf{0.00} $\pm$ 0.00
,FRPD-DPP-LS,1.52 $\pm$ 0.42,\textbf{4.50} $\pm$ 1.38,\textbf{0.00} $\pm$ 0.00
Adult,FACE,\textbf{1.33} $\pm$ 0.36,1.57 $\pm$ 0.83,0.12 $\pm$ 0.28
,FRPD-QUAD,1.99 $\pm$ 0.59,3.80 $\pm$ 0.50,\textbf{0.01} $\pm$ 0.02
,FRPD-DPP-GR,2.00 $\pm$ 0.52,\textbf{3.87} $\pm$ 0.43,0.05 $\pm$ 0.13
,FRPD-DPP-LS,2.00 $\pm$ 0.52,\textbf{3.87} $\pm$ 0.43,0.05 $\pm$ 0.13
\end{filecontents*}
\pgfplotstabletypeset[
        col sep=comma,
        string type,
        every head row/.style={before row=\toprule,after row=\midrule},
        every row no 3/.style={after row=\midrule},
        every row no 7/.style={after row=\midrule},
        every row no 11/.style={after row=\midrule},
        every last row/.style={after row=\bottomrule},
        columns/dataset/.style={column name=Dataset, column type={l}},
        columns/method/.style={column name=Methods, column type={l}},
        columns/cost/.style={column name=Shortest Path Cost, column type={c}},
        columns/lev/.style={column name=Path Diversity, column type={c}},
        columns/jac/.style={column name=Path Anti Diversity, column type={c}},
    ]{mlp_diverse_path_updated.csv}
\end{table*}

\subsection{Comparisons of Prototype Selection Algorithms}

\begin{figure}[H]
    \centering
      \includegraphics[width=1.0\linewidth]{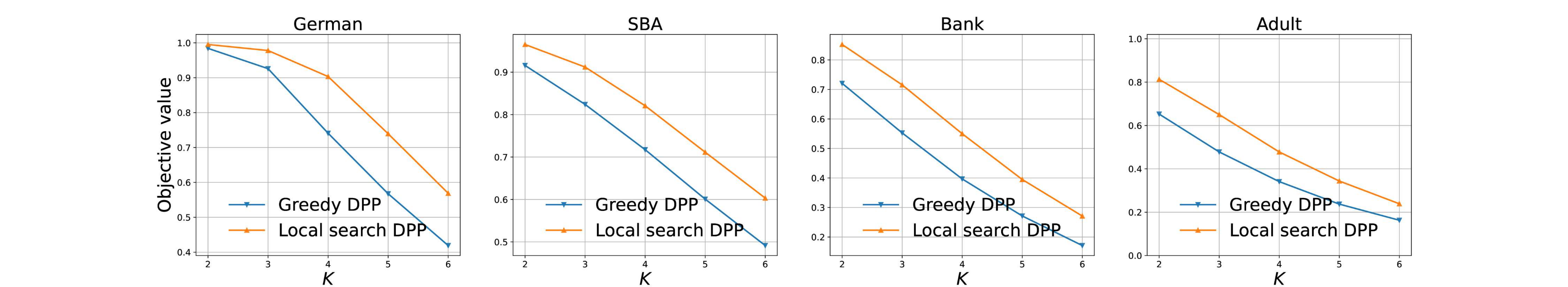}
    \caption{Average objective value (higher is better) of Greedy solution and Greedy+Local search solution with a different number of prototypes $K$, the average is taken over 100 randomly chosen input $x_0$ from the test set.}
    \label{fig:dpp_obj}
\end{figure}

\textbf{Greedy and Local search DPP.} Figure~\ref{fig:dpp_obj} presents the additional improvement of the solution quality, measured by the determinant value in the objective function of~\eqref{eq:det}, that is generated by the local neighborhood search. We observe that across all datasets, the local search constantly improves the quality of the greedy heuristics.

\textbf{Greedy and Eigen-Approximate Binary Quadratic Program.} We also implement a greedy heuristic search to solve problem~\eqref{eq:cosine2}. 
This greedy algorithm proceeds by finding iteratively for each incumbent set of prototypes $z$ an index $j$ by
\[ 
    j = \argmin_{i: z_i = 0}~\vartheta (z_i')^\top S z_i' + (1-\vartheta) d^\top z_i', \quad \text{where} \quad z_i' = z \vee \bar e_i
\]
with $\vee$ denoting the element-wise maximum between two vectors and $e_i$ is the vector of zeros with the $i$-th element being one. The algorithm then adds $j$ to the set of prototypes until reaching the cardinality constraint, i.e., when $\|z\|_0 = K$. In doing so, each $j$ guarantees to maximize the marginal gain to the incumbent set.

We compare the solution of the greedy method against our approximate solution where the quality of the solution is measured by the objective value of problem~\eqref{eq:cosine2}.
\begin{figure}[H]
    \centering
      \includegraphics[width=1.0\linewidth]{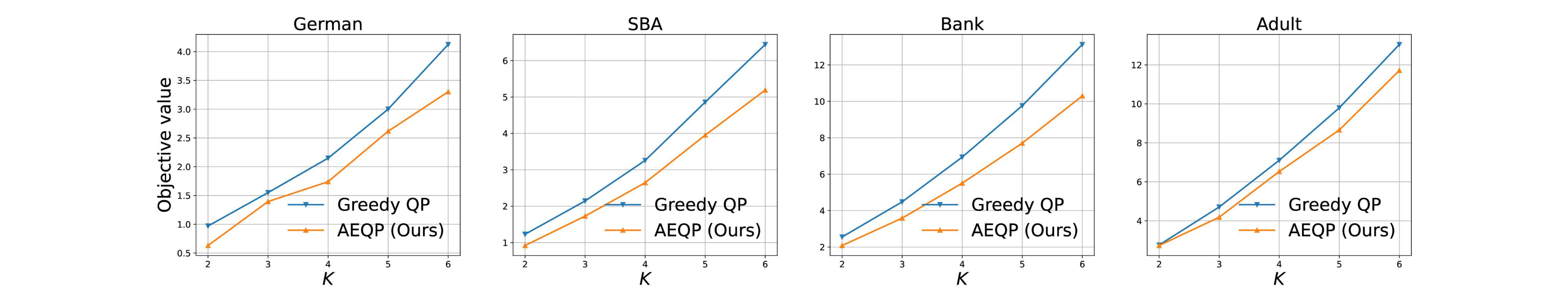}
    \caption{Average objective value (lower is better) of Greedy solution and Eigen-Approximate Binary Quadratic Program solution with a different number of prototypes $K$, the average is taken over 100 randomly chosen input $x_0$ from the test set.}
    \label{fig:qp_obj}
\end{figure}

\textbf{Comparison of run time.} We compare the run time of our three approaches (FRPD-QUAD, FRPD-DPP-GR, FRPD-DPP-LS) on synthetic datasets. We generate 2-dimensional data samples by sampling uniformly in a rectangle $x = (x_1, x_2) \in [-2, 4] \times [-2, 7]$ with the following binary labelling function $f$:
\[
    f(x) = \left\{
            \begin{array}{cl}
                1 & \mathrm{if} \quad x_2 \ge 1 + x_1 + 2 x_1^2 + x_1^3 - x_1^4, \\
                0 & \mathrm{otherwise},
            \end{array}
        \right.
\]

\begin{figure}[H]
    \centering
      \includegraphics[width=0.4\linewidth]{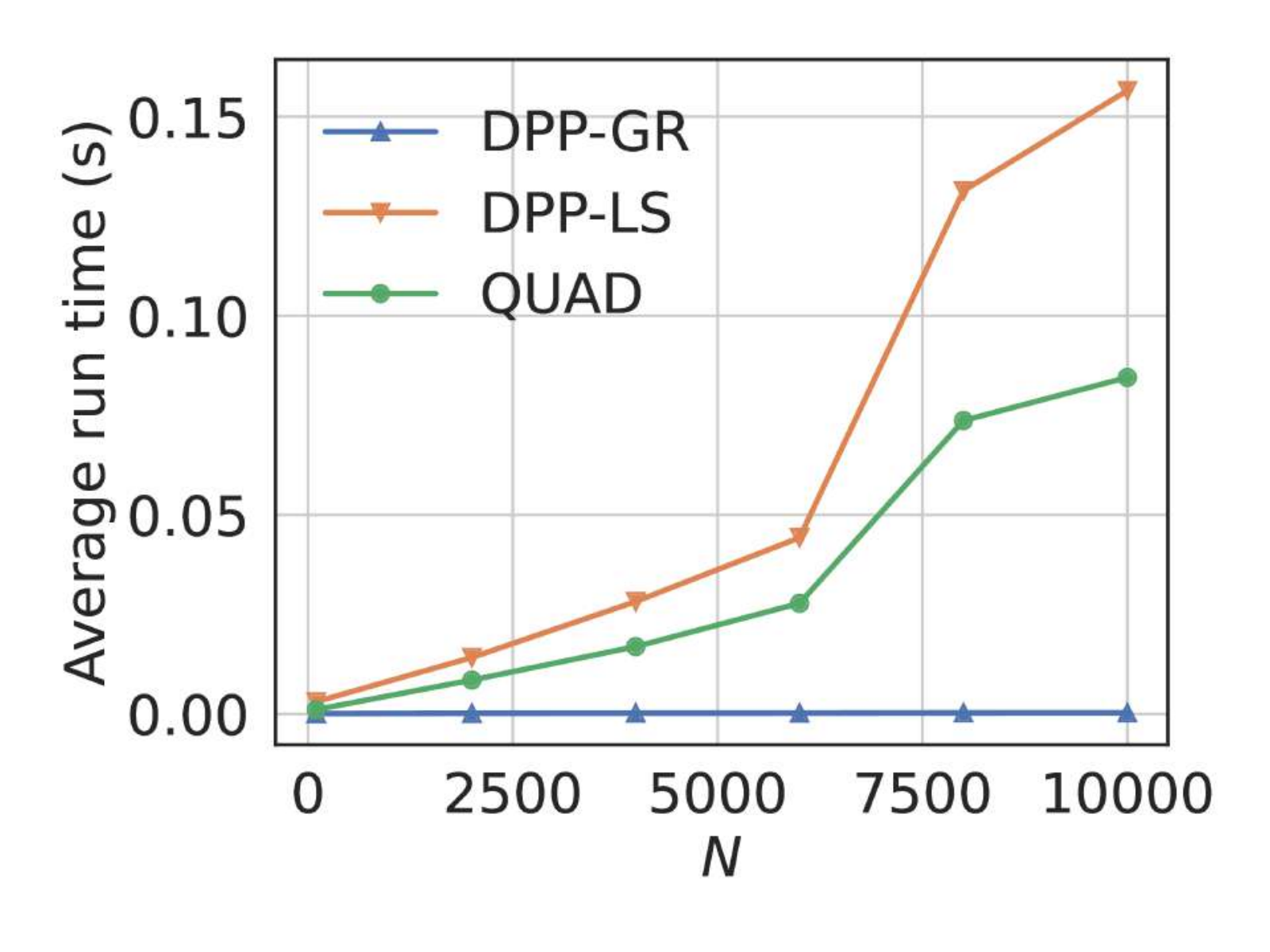}
    \caption{Average run time(s) of three different approaches with different sample sizes $N$ over 5 independent replications for FRPD-QUAD (QUAD), FRPD-DPP-GR (DPP-GR) and FRPD-DPP-LS (DPP-LS).}
    \label{fig:time}
    \end{figure}

First, we synthesize $N$ 2-dimensional data samples for each value $N = 100, \ldots, 10000$. For each negatively predicted instance, we find $K=3$ prototypes in the positively predicted class. The process is repeated independently five times, then we take the average run time of our three methods and report the results in Figure~\ref{fig:time}. These results indicate that the Greedy DPP has the smallest run time. The run time of the Local search DPP increases significantly when the number of samples increases. Furthermore, these findings demonstrate that our FRPD-QUAD approach is appropriate for the Anti-Diversity measure, as it has a comparable run time to FRPD-DPP-GR and the lowest Anti-Diversity in all the experiments.

\begin{figure}[H]
    \centering
      \includegraphics[width=0.4\linewidth]{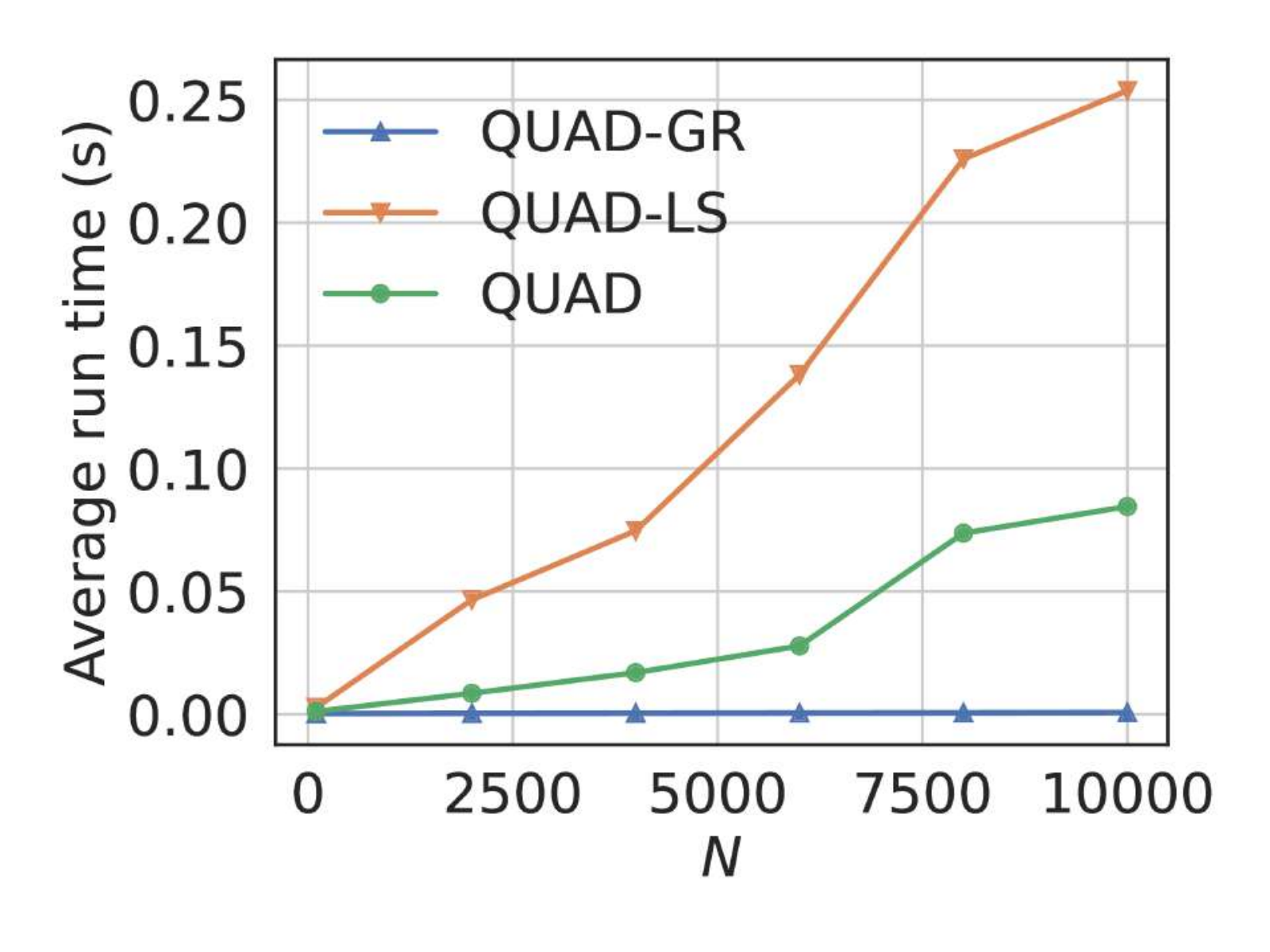}
    \caption{Average run time(s) of three different approaches with different sample sizes $N$ over 5 independent replications for FRPD-QUAD (QUAD), QUAD-Greedy (QUAD-GR), and QUAD-Local search (QUAD-LS).}
    \label{fig:time_qp}
    \end{figure}

Additionally, we evaluate the run time of FRPD-QUAD (QUAD), QUAD-Greedy (QUAD-GR), and QUAD-Local search (QUAD-LS) and report the results in Figure~\ref{fig:time_qp}. These results demonstrate that the Greedy method has the smallest run time. The run time of the Local search quadratic increases significantly when the number of samples increases.

\section{Background on Determinant Point Processes} \label{sec:dpp}
We first provide a definition of a DPP.
\begin{definition}[Determinantal point process]
    Given a set $[N] \Let \{1, \ldots, N\}$ containing $N$ items, a DPP defines a probability measure over all subsets of $[N]$ that is parametrized by a matrix $P \in \PSD^N$, $0 \preceq P \preceq I$. If $\tilde R$ is a random subset drawn according to this DPP, then for every subset $J \subseteq [N]$,
    \[
        \mathrm{Probability}_{ P}(J \subseteq \tilde R) = \det( P_J),
    \]
    where $P_J$ is a submatrix of $P$ obtained by restricting to rows and columns indexed in $J$.
\end{definition}

The bound conditions $0~\preceq~ P~\preceq~ I$ ensure that all principal minors of $P$ are nonnegative and smaller than one, which is sufficient to define a proper probability distribution. The matrix $P$ is called the marginal kernel. If $J = \{i\}$ is a singleton set, then $\mathrm{Probability}_{P}(i \in \tilde R) =  P_{ii}$. If $J = \{i, j\}$ has two items, then
\[
    \mathrm{Probability}_{P} (i, j \in \tilde R) = \det \left( \begin{bmatrix}
         P_{ii} &  P_{ij} \\
         P_{ji} &  P_{jj}
    \end{bmatrix}\right)
    =  P_{ii}  P_{jj} -  P_{ij}  P_{ji}.
\]
The term $- P_{ij}  P_{ji}$ captures the relationship between two items $i$ and $j$. Since $ P$ is symmetric, we have $- P_{ij}  P_{ji} = - P_{ij}^2 \leq 0$; thus DPPs are able to model the negative correlations between items. The larger value of $P_{ij}$, the lower probability of items $i$ and $j$ co-occurs. If $P_{ij} = 0$, there is no interaction between the two items.

In practice, it is difficult to deal with a generic correlation kernel $P$. Therefore, $L$-ensemble, a subclass of DPPs, is often used instead since it provides several simpler formulas. For a positive semidefinite matrix $L \in \PSD^N$, an $L$-ensemble DPP specifies the atomic probabilities for every possible instantiation of $\tilde R$ via
\[
    \mathrm{Probability}_{ L}(\tilde R = J) \propto \det( L_J) \qquad \forall J \subseteq [N].
\]
The normalization constant can be computed as $\sum_{J \subseteq [N]} \det( L_J) = \det( L +  I)$. Here, a positive semidefinite matrix $ L$ is admissible to define a DPP with a marginal kernel matrix $P =  I -  ( L +  I)^{-1}$. Typically, the matrix $L$ is chosen as a similarity matrix, in which the diagonal element $ L_{ii}$ represents the quality of item $i$ while the off-diagonal element $L_{ij}$ is the similarity measure between $i$ and $j$. 

\section{Local Search for the Maximum A Posteriori (MAP) Inference} \label{sec:ls}

We describe in this section a neighborhood heuristic that can be employed to improve the greedy algorithm of the MAP inference problem in Section~\ref{sec:det}.

Let $z \in \{0, 1\}^N$, $\| z \|_0 = K$ be the current solution of the local search algorithm. We define the local move for the solution $z$ by removing the element $j^{-}$  that has the \textit{smallest} marginal decrease in the objective value and then adding the element $j^{+}$ that engenders the \textit{largest} marginal increase in the objective value. In detail, we define
\[
    j^{-} \Let \argmin_{i: z_i=1} \; \log \det (L_{z \wedge \bar{e}_i}) - \log \det (L_z)
    \]
and compute
   \[
    j^{+} \Let \argmax_{i: (z \wedge \bar{e}_{j^-})_i=0} \; \log \det (L_{z \wedge \bar{e}_{j^-} \vee e_i}) - \log \det (L_{z \wedge \bar{e}_{j^-}}),
\]
where $\bar{e}_i$ is the vector of ones with the $i$-th element being zero. The operator $\wedge$ is the bitwise OR and $\vee$ is the bitwise AND operator. Computing $j^-$ costs $\mc O(K^4)$ in running time while we can find $j^+$ in $\mc O(NK^2)$ time using similar procedure as in~\cite{ref:chen2018fast}. For $K \ll N$, the total complexity for this local move is $\mc O(NK^2)$.

For the local search procedure, we first initialize the solution using the output of the greedy algorithm and then iteratively perform the local moves. This procedure terminates when there is no neighboring solution that can improve the objective value.

\section{Proof} \label{sec:app-proof}

We provide here the proof of Lemma~\ref{lemma:minmax} that is omitted in the main text.

\begin{proof}[Proof of Lemma~\ref{lemma:minmax}]
    If we add the auxiliary variables $h_m$ and the resulting constraints $h_m = v_m^\top z$, then problem~\eqref{eq:cosineK} becomes
    \[
        \begin{array}{cl}
            \min & (1 - \vartheta) d^\top z + \vartheta \sum_{m=1}^M \sigma_m h_m^2 \\
            \st & h \in \R^M,~z \in \{0, 1\}^N \\
            & \|z \|_0 = K, \quad h_m = v_m^\top z \quad \forall m = 1, \ldots, M.
        \end{array}
    \]
    This optimization problem can be decomposed into a two-layer problem of the form
    \[
        \begin{array}{ccl}
            \Min{z \in \{0, 1\}^N, \|z\|_0 = K} & \min & (1 - \vartheta) d^\top z + \ds \vartheta \sum_{m=1}^M \sigma_m h_m^2 \\
            &\st & h \in \R^M,~h_m = v_m^\top z \quad \forall m = 1, \ldots, M.
        \end{array}
    \]
    For any feasible solution $z$, the inner minimization problem over $h$ is a convex optimization problem. By strong duality, we have the equivalent problem
    \[
        \Min{\substack{z \in \{0, 1\}^N \\ \|z\|_0 = K}} ~ \Max{\gamma \in \R^M} ~ \Min{h \in \R^M}~\ds (1 - \vartheta) d^\top z + \vartheta \sum_{m=1}^M \sigma_m h_m^2 + 2\sum_{m=1}^M \gamma_m (h_m - v_m^\top z)
    \]
    The optimization problem in $h$ is separable, and because $\sigma_m > 0$, the optimal solution is
    \[
        h_m\opt = - \frac{\gamma_m}{\vartheta \sigma_m} \quad \forall m = 1, \ldots, M.
    \]
    By replacing this optimal solution with the objective function, we thus obtain the equivalent problem. This completes the proof.
\end{proof}
\end{document}